\documentclass[conference,compsoc]{IEEEtran}
\usepackage{graphicx}
\usepackage{subfigure}
\usepackage{float}
\usepackage{verbatim}
\usepackage{multirow}

\ifCLASSOPTIONcompsoc
  \usepackage[nocompress]{cite}
\else
  \usepackage{cite}
\fi
\ifCLASSINFOpdf
\else
\fi

\usepackage{xcolor}

\begin{document}
\title{Performance and Power Modeling and Prediction Using MuMMI and Ten Machine Learning Methods}
%

\author{\IEEEauthorblockN{Xingfu Wu, Valerie Taylor}
\IEEEauthorblockA{Argonne National Laboratory\\ University of Chicago\\
Email: \{xingfu.wu, vtaylor\}@anl.gov}
\and
\IEEEauthorblockN{Zhiling Lan}
\IEEEauthorblockA{Department of Computer Science \\ Illinois Institute of Technology\\
            Email: lan@iit.edu}
}

\maketitle

\thispagestyle{plain}
\pagestyle{plain}

\begin{abstract}

In this paper, we use modeling and prediction tool MuMMI (Multiple Metrics Modeling Infrastructure) and ten machine learning methods to model and predict performance and power and compare their prediction error rates.  We use a fault-tolerant linear algebra code and a fault-tolerant heat distribution code to conduct our modeling and prediction study on the Cray XC40 Theta and IBM BG/Q Mira at Argonne National Laboratory and the Intel Haswell cluster Shepard at Sandia National Laboratories. Our experiment results show that the prediction error rates in performance and power using MuMMI are less than 10\% for most cases. Based on the models for runtime, node power, CPU power, and memory power, we identify the most significant performance counters for potential optimization efforts associated with the application characteristics and the target architectures, and we predict theoretical outcomes of the potential optimizations. When we compare the prediction accuracy using MuMMI with that using 10 machine learning methods, we observe that MuMMI not only results in more accurate prediction in both performance and power but also presents how performance counters impact the performance and power models. This provides some insights about how to fine-tune the applications and/or systems for energy efficiency. 

\end{abstract}


\section{Introduction}
\label{intro}
 Energy-efficient scientific applications require insight into how HPC system features impact the applications' power and performance. This insight can result from the development of performance and power models. Dense matrix factorizations, such as LU, Cholesky, and QR, are widely used for scientific applications that require solving systems of linear equations, eigenvalues, and linear least squares problems \cite{DB12} \cite{BH15}. 
 Such real-world scientific applications take a long time to execute on supercomputers, thereby relying on resilience techniques to successfully finish the long executions because of software and hardware failures. 
 While reducing execution time is still a major objective for high-performance computing, future HPC systems and applications will have additional power and resilience requirements that represent a multidimensional tuning challenge. To embrace these key challenges, we must understand the complicated tradeoffs among runtime, power, and resilience. In this paper we explore performance and power modeling and prediction of an algorithm-based fault-tolerant linear algebra code (FTLA)  \cite{FTLA} and a fault-tolerant heat distribution code (HDC) \cite{8} using MuMMI (Multiple Metrics Modeling Infrastructure) \cite{19} \cite{WT16} and ten machine learning methods  \cite{KJ13} \cite{CAR}.
 
       In this work, we use FTLA and HDC to conduct our experiments on the Cray XC40 Theta \cite{Theta} and IBM BG/Q Mira \cite{MIRA} at Argonne National Laboratory  and on the Intel Haswell cluster Shepard \cite{SDA} at Sandia National Laboratories. We analyze FTLA's performance and power characteristics and use MuMMI and ten machine learning methods to model, predict and compare performance and power of FTLA and HDC. MuMMI \cite{19} is a tool infrastructure that facilitates systematic measurement, modeling, and prediction of performance and power consumption, and performance-power tradeoffs and optimization for parallel systems. The ten machine learning methods from the R caret package \cite{CAR} \cite{KJ13} are Random Forests \cite{LW18}, Gaussian Process with Radial Basis Function \cite{KSH}, eXtreme Gradient Boosting \cite{CH19}, Stochastic gradient boosting \cite{GB19}, Cubist \cite{KW20}, Ridge Regression \cite{ZH18}, k-Nearest Neighbors \cite{KJ13}, Support Vector Machines with Linear Kernel \cite{KSH}, Conditional Inference Tree \cite{HH20}, and Multivariate Adaptive Regression Spline \cite{MS19}. 
       
       Our experiment results show that the prediction error rates in performance and power using MuMMI are less than 10\% for most cases. Based on the models for runtime, node power, CPU power, and memory power, we identify the most significant performance counters for potential optimization efforts associated with the application characteristics and the target architectures, and we predict theoretical outcomes of the potential optimizations. When we compare the prediction accuracy using MuMMI with that using 10 machine learning methods, we observe that MuMMI results in more accurate prediction in both performance and power. 

    The remainder of this paper is organized as follows.  Section 2 discusses the FTLA and HDC. Section 3 briefly describes three architectures and their power profiling tools.  Section 4 presents performance and power characteristics, modeling and prediction of FTLA using MuMMI. Section 5 discusses the modeling and prediction of FTLA and HDC using 10 machine learning methods and compares them with MuMMI. Section 6 summarizes this work. Notice that we use the formula $(prediction - baseline)/baseline *100\%$ to calculate the prediction error rate in this paper.

\section{Fault-Tolerant Applications: FTLA and HDC}

A number of resilience methods have been developed for preventing or mitigating failure impact. 
 Existing resilience strategies can be broadly classified into four approaches: checkpoint based, redundancy based, proactive methods, and algorithm based. Checkpoint/restart is a long-standing fault tolerance technique to alleviate the impact of system failures, in which the applications save their state periodically, then restart from the last saved checkpoint in the event of a failure. Multilevel checkpointing is the state-of-the-art design of checkpointing, focusing on reducing checkpoint overhead to improve checkpoint efficiency. Such checkpointing libraries include FTI (Fault Tolerance Interface) \cite{2} \cite{8}, SCR (Scalable Checkpoint/Restart) \cite{17} \cite{12}, VeloC \cite{Vlc}, and diskless checkpointing \cite{15}. Redundancy approaches improve resilience by replicating data or computation \cite{5} \cite{6} \cite{7}.  Proactive methods take preventive actions before failures, such as software rejuvenation and process or object migration \cite{13}. 
 Algorithm-based fault tolerance (ABFT) methods maintain consistency of the recovery data by applying appropriate mathematical operations on both the original and recovery data, and they adapt the algorithm so that the application dataset can be recovered at any moment \cite{HA84} \cite{AL88} \cite{4} \cite{BD09} \cite{BH15}. ABFT was applied to High Performance Linpack (HPL) \cite{DK11}, to  Cholesky factorization \cite{HC10}, and to  LU and QR factorizations \cite{DB12} \cite{DL11}  \cite{BH15}. The FTLA  \cite{FTLA} \cite{BH15} in particular was developed as an extension to ScaLAPACK \cite{SLK} that tolerates and recovers from fail-stop failures, which 
  is defined as a process that completely stops responding, triggering the loss of a critical part of the global application state, and halts the application execution. 

\begin{figure}
\center
 \includegraphics[height=1.6in, width=1.4in]{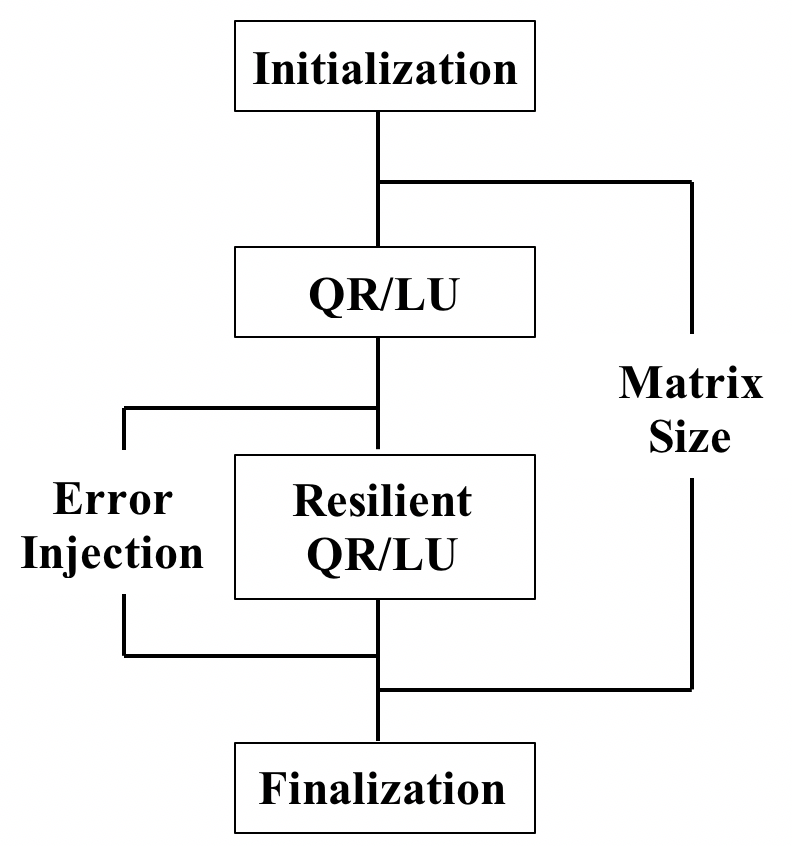}
 \caption{Control flow of  the ftla-rSC13 code}
\label{fig:1}       
\end{figure}

Matrix QR factorization decomposes a matrix A into a product A = QR, where Q is an orthogonal matrix and R is an upper triangular matrix. The  code ftla-rSC13 \cite{FTLA} consists of two main components: one QR operation followed by a resilient QR (RQR) operation, where the RQR performs one QR, checkpointing, and repairing a failure until completing without a failure as shown in Figure \ref{fig:1}. The structure of the block QR and LU is identical.  We focus on the QR in this paper. The main loop is associated with the matrix sizes. For each matrix size, it performs one QR followed by one small loop. The small loop size is the number of error injections. For each error injection, it performs one RQR.

We remove all segments for error injections from ftla-rSC13 to create another code called $ftla$. The main loop is associated with the matrix sizes. For each matrix size, it performs one QR followed by one RQR. The RQR performs one QR and checkpointing. Then we remove the checkpointing segments from $ftla$ to get a code called $la$, which is similar to ScaLAPACK QR. In this paper, we use the three codes ftla-rSC13, $ftla$, and $la$ to conduct our experiments. They are strong scaling.

The other application used in this paper is an FTI version of MPI heat distribution benchmark code (HDC) \cite{8}, which computes the heat distribution over time based on a set of initial heat sources. FTI \cite{2} leverages local storage, along with data replication and erasure codes, to provide several levels of reliability and performance. It provides four-levels checkpointing: local write (L1), Partner copy (L2), Reed-Solomon coding (L3), and PFS write (L4).The four checkpointing levels correspond to coping with the four types of failures: no hardware failure (software failure), single-node failure, multiple-node failure,  and all other failures the lower levels cannot take care of, respectively.  The checkpointing file size is 2 MB per MPI process. HDC is a compute-intensive, weak scaling. 

 While fault tolerance methods and power-capping techniques continue to evolve, tradeoffs among execution time, power efficiency, and resilience strategies are still not well understood. Fault tolerance studies focus mainly on the tradeoffs between execution time,
fault tolerance overhead, and resiliency, whereas most power management studies focus on the tradeoffs between execution time and power. Understanding the tradeoffs among all  these factors is crucial because future HPC systems will be built under both reliability and power constraints. The previous work \cite{WT18} presented an empirical study evaluating the runtime and power requirements of multilevel checkpointing MPI applications using FTI on four different parallel architectures. Recent research has focused on a theoretical analysis of energy and runtime for fault tolerance protocols  \cite{AB13} \cite{MS14} \cite{1} \cite{EG12} \cite{EG13} \cite{TK14}. In this paper, we use the FTLA and HDC to conduct our modeling and prediction study.

\section{System Architectures and Environments}
\label{sec:3}

We conduct our experiments on three parallel systems with different architectures: the Cray XC40 Theta \cite{Theta} and IBM BG/Q Mira \cite{MIRA} at Argonne National Laboratory and the Intel Haswell cluster Shepard \cite{SDA} at Sandia National Laboratories.  Details about each system are given in Table~\ref{tab:1}. Each Cray XC40 node has 64 compute cores: one Intel Phi Knights Landing (KNL) 7230 with the thermal design power (TDP) of 215 W, 32 MB of L2 cache, 16 GB of high-bandwidth in-package memory (MCDRAM), 192 GB of DDR4 RAM, and a 128 GB SSD. Each BG/Q node has 16 compute cores---one BG/Q PowerPC A2 1.6 GHz chip with the TDP of 55 W \cite{2}) and shared L2 cache of 32 MB and 16 GB of memory.  Each Haswell node has 32 compute cores---two Xeon E5-2698 V3  2.3 GHz chips with the TDP of 135 W per chip and shared L3 cache of 40 MB and 128 GB of memory. 

Theta uses the Cray Aries dragonfly network with user access to a Lustre parallel file system with 10 PB of capacity and 210 GB/s bandwidth \cite{Theta}. Mira uses a 5D torus network with user access to a GPFS file system \cite{MIRA}. Shepard uses a Mellanox fourteen data rate InfiniBand network with a regular NFS file system \cite{SDA}.

Several vendor-specific power management tools exist, such as Cray's CapMC and out-of-band and in-band power monitoring capabilities \cite{MR16}, IBM EMON API on BG/Q \cite{2}, Intel RAPL \cite{16}, and NVIDIA's power management library \cite{14}. In this work, we use  simplified PoLiMEr \cite{MV17} to measure power consumption for the node, CPU, and memory at the node level on Theta; we use MonEQ \cite{18} to collect power profiling data on Mira; and we use PowerInsight \cite{10} to measure the power consumption for the node, CPU, memory, and hard disk at the node level on Shepard.

\begin{table}
\center
\caption{Specifications of three different architectures}
\begin{tabular}{c}
  \includegraphics[height=2.2in, width=3.3in]{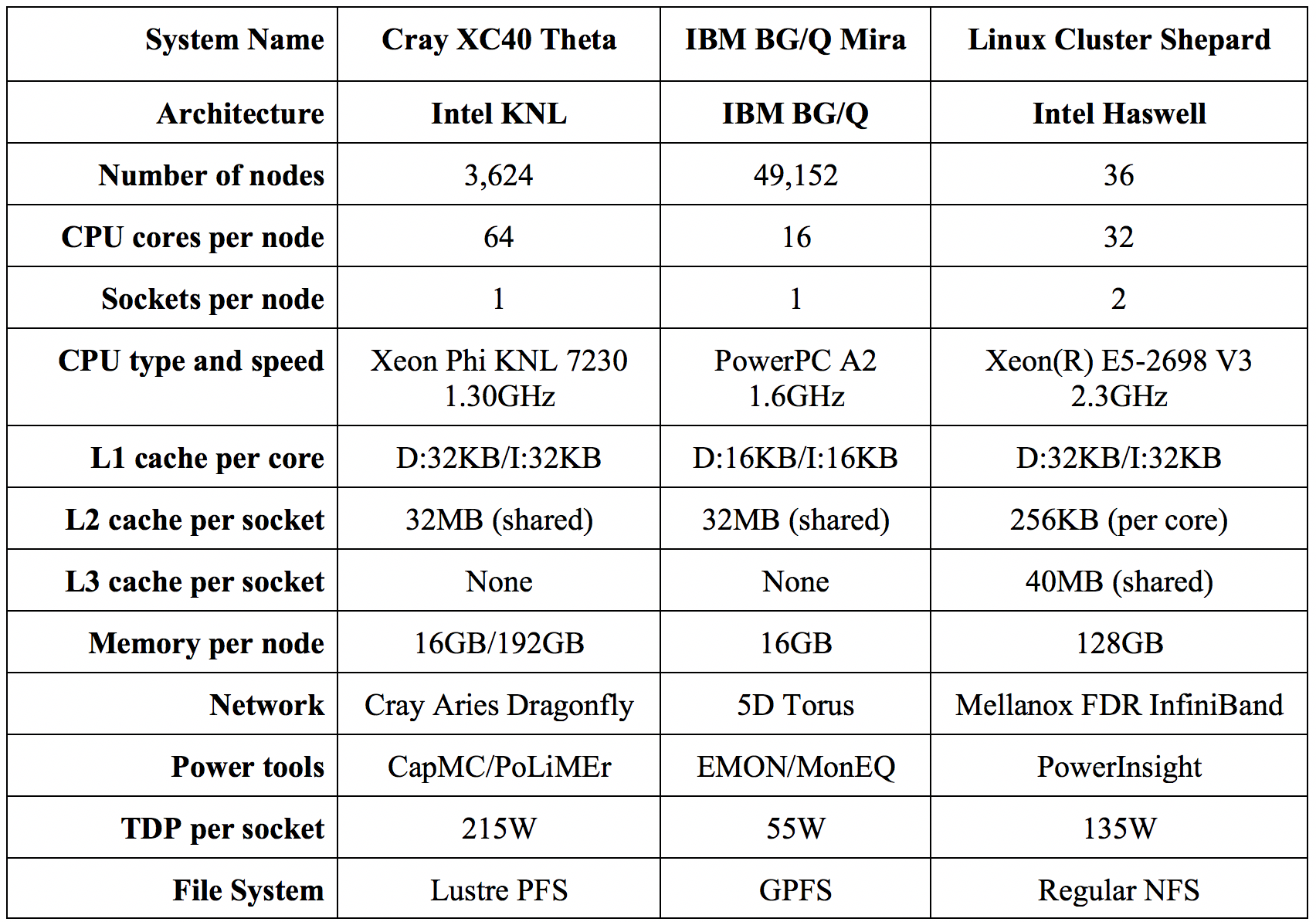}
  \end{tabular}
\label{tab:1}       
\end{table}  

PoLiMEr uses Cray's CapMC to obtain power and energy measurements of the node, CPU, and memory on Theta. The power-sampling rate used is approximately 2 samples per second (default). On Mira, EMON API \cite{2} provides 7 power domains to measure the power consumption for the node, CPU, memory, and network at the node-card level. The power-sampling rate used is approximately 2 samples per second (default). Each node-card consists of 32 nodes. To obtain the power consumption at the node level, we calculate the average power by dividing by 32. So we conduct our experiments on multiple node-cards to obtain the power-profiling data. PowerInsight provides the measurement for 10 power rails for CPU, memory, disk, and motherboard on the Intel Haswell system Shepard. The power-sampling rate used is 1 sample per second (default).

\section{Modeling and Prediction Using MuMMI}

In this section, we use the three codes ftla-rSC13, ftla, and la with matrix sizes from 6,000 to 20,000 with a stride of 2000 and a block size of 100 to conduct our FTLA experiments with a maximum of error injections of 5 on Cray XC40 Theta, IBM BG/Q Mira, and Intel Haswell Shepard. We analyze their performance and power characteristics, and use performance counter-based modeling tool MuMMI \cite{19} \cite{WT16} to model performance and power and to predict theoretical outcomes for the potential optimizations.

We use MuMMI with support of PoLiMEr, MonEQ and PowerInsight to instrument these codes to collect performance data, power data, and performance counters on Cray XC40 Theta, IBM BG/Q Mira, and Intel Haswell Shepard.  We use the same default compiler options from the FTLA code ftla-rSC13 to compile the codes.  For a given application run, we execute the application 14 times on each system to ensure the consistency of the results while collecting different sets of performance counters with a total of 40 performance counters for performance and power modeling. We found that the variation of the application runtime is very small (less than 1\%), so we use the performance metrics corresponding to the smallest runtime for our work.

\subsection{Cray XC40 Theta}

Each XC40 node \cite{Theta} has one Intel KNL, which brings in an on-package memory called Multi-Channel DRAM (MCDRAM) in addition to the traditional DDR4 RAM. MCDRAM has a high bandwidth (4 times more than DDR4 RAM) and low capacity (16 GB) memory. MCDRAM can be configured as a shared L3 cache (cache mode) or as a distinct NUMA node memory (flat mode). With the different memory modes by which the system can be booted, it becomes  challenging from a software perspective to understand the best mode  for an application. We use the codes $la$ and $ftla$ to investigate the performance and power impacts under the cache or flat mode use of MCDRAM. 

\begin{figure}
\center
 \includegraphics[height=1.8in, width=3.3in]{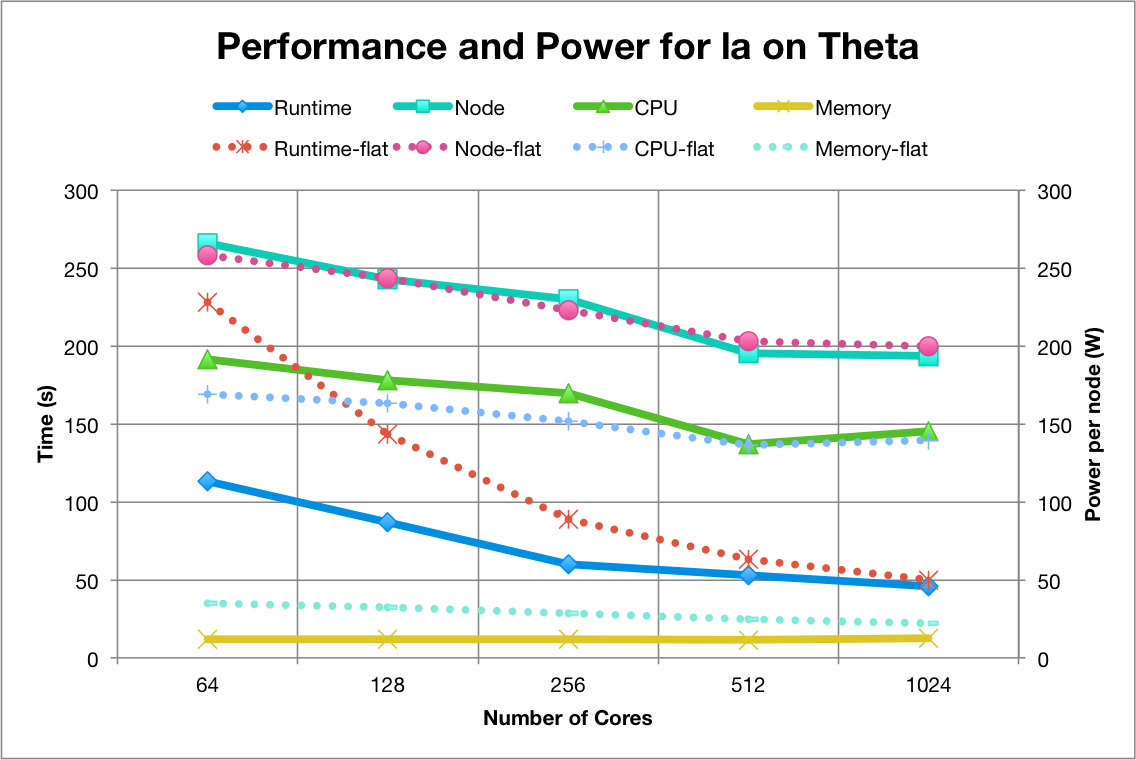}
 \caption{Performance and power comparison for la on Theta}
\label{fig:14}       
\end{figure}

\begin{figure}
\center
 \includegraphics[height=1.8in, width=3.3in]{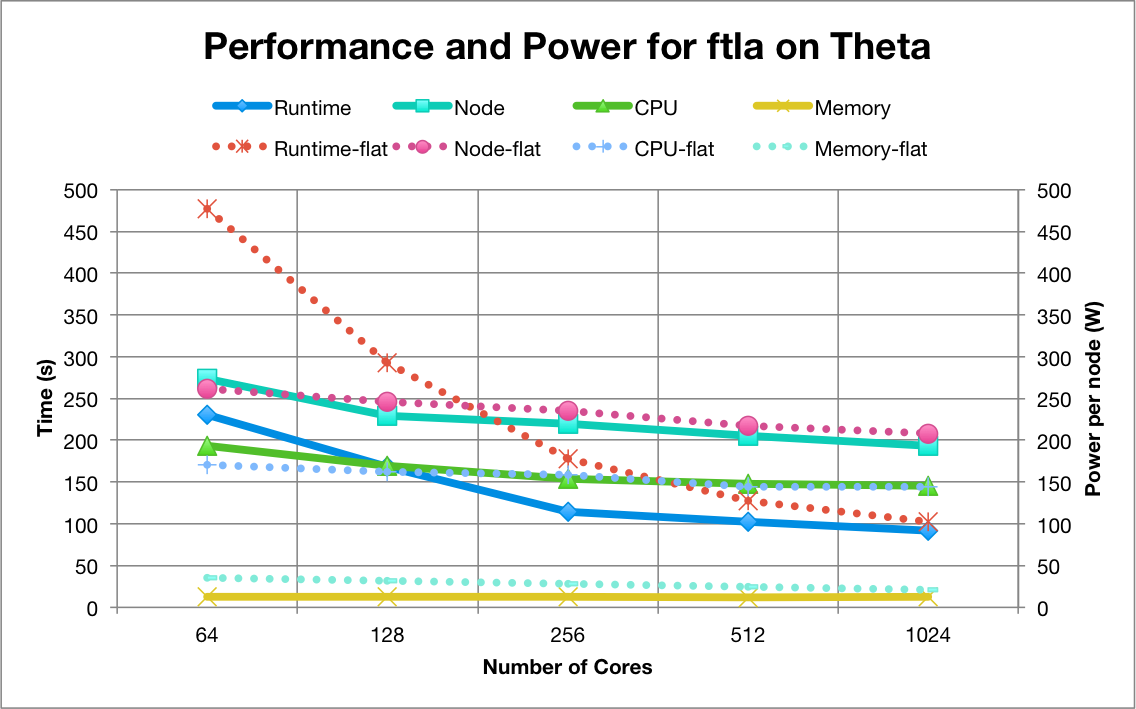}
 \caption{Performance and power comparison for ftla on Theta}
\label{fig:15}       
\end{figure}

 Figure \ref{fig:14} presents the performance and power comparison of $la$ using the two memory modes on Theta, where the terms with \textbf{-flat} stand for using the flat mode and the terms without \textbf{-flat} stand for using the cache mode. The advantage of using MCDRAM as cache is that an application may run entirely in MCDRAM so that the application performance may be improved significantly. We find that the application runtime using the cache mode is almost half of the runtime using the flat mode on 64 cores. Both runtimes are close, however, increasing the number of cores to 1,024 because the application is strong scaling, the amount of workload per core decreases by 16 times. These results indicate that, to take advantage of MCDRAM requires a large amount of workload per core. From the figure, we also observe that both node power consumptions are close. The CPU power for using the cache mode is higher, but the memory power for using the cache mode is much lower. Overall, using the cache mode results in lower energy consumption for these cases. 
We find the same trend for the code ftla shown in Figure \ref{fig:15}. Therefore, in the remainder of this section, we use the cache mode to conduct our experiments on Theta.

\begin{figure}
\center
 \includegraphics[height=1.8in, width=3.3in]{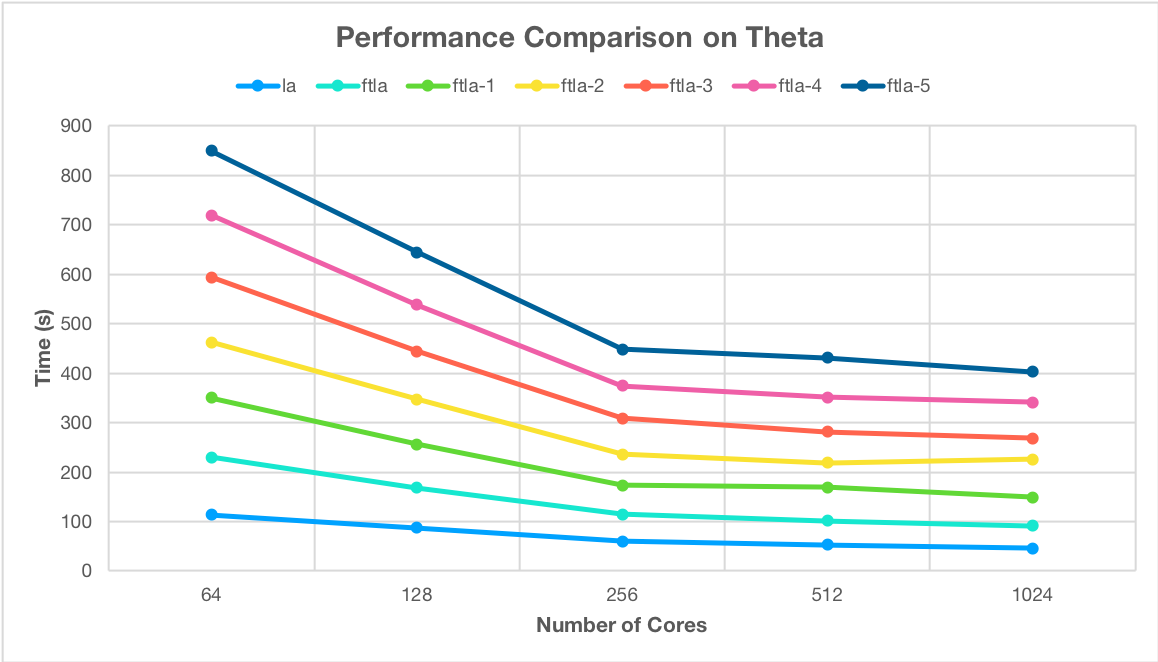}
 \caption{Performance comparison on Theta}
\label{fig:9}       
\end{figure}

Figure \ref{fig:9} presents a performance comparison of the three codes on Theta, where \textbf{ftla-1} stands for the code ftla-rSC13 with one error injection. We observe a proportional increase in application runtime with increasing numbers of error injections on up to 1,024 cores because of the proportional increase in the number of error injections. 

 \begin{figure}
\center
 \includegraphics[height=1.6in, width=3.3in]{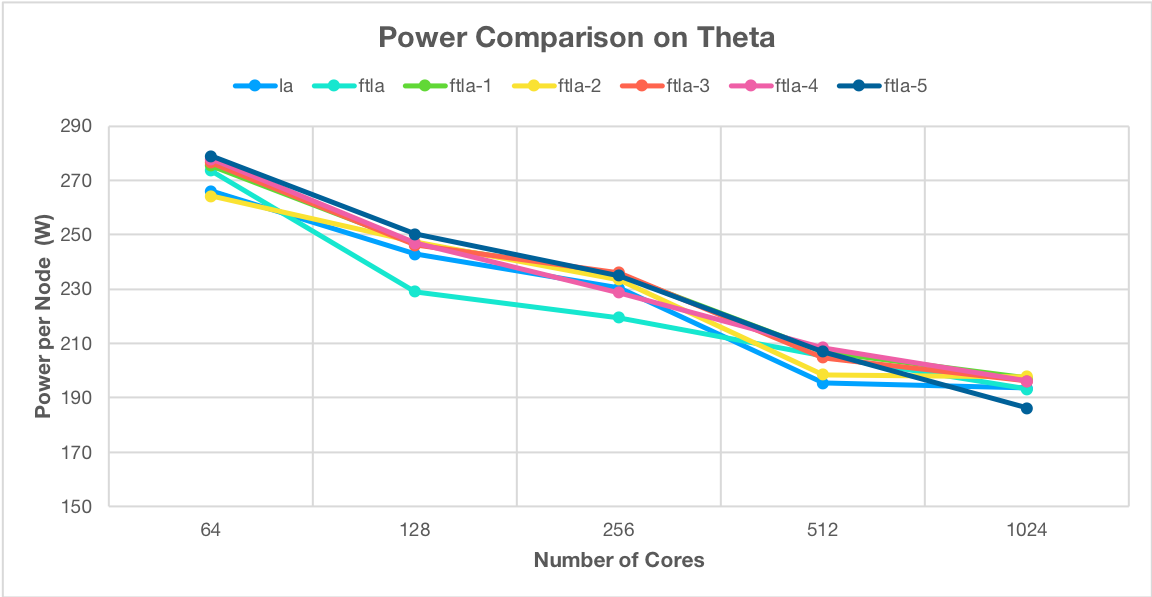}
 \caption{Power comparison on Theta}
\label{fig:12}       
\end{figure} 

\begin{figure}
\center
 \includegraphics[height=1.6in, width=3.3in]{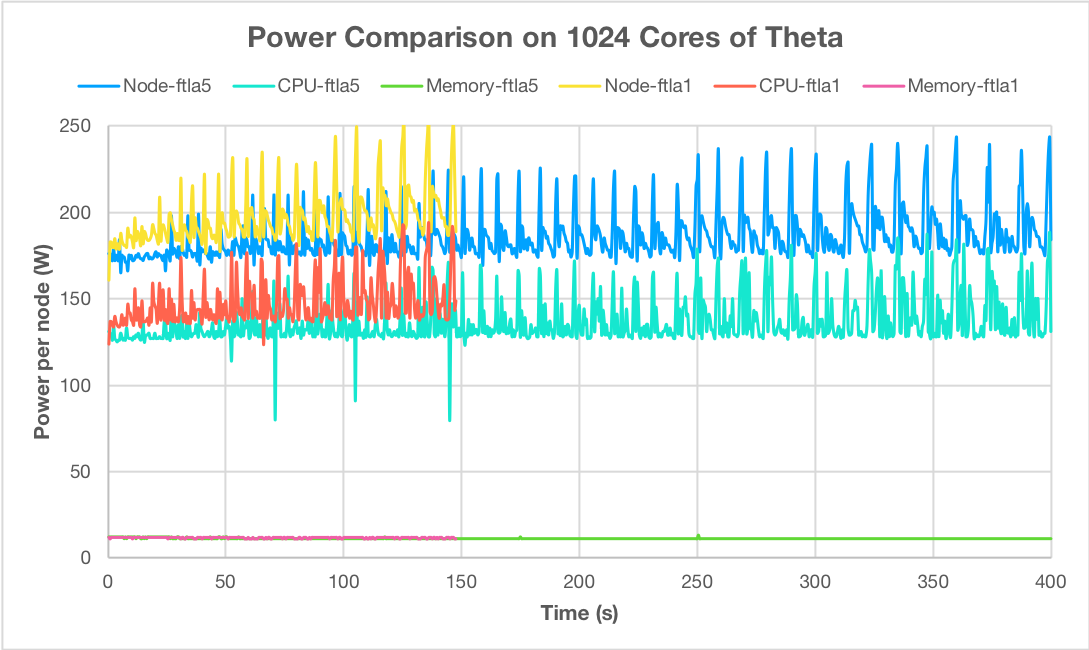}
 \caption{Power over time on 1024 cores of Theta}
\label{fig:31}       
\end{figure}

Figure \ref{fig:12} shows the average node power consumption on Theta. The node power consumption decreases with increasing numbers of cores because of the strong scaling and dynamic power management support. 
Further, we compare power over time for the FTLA with one error injection and five error injections on 1,024 cores in Figure \ref{fig:31}. 
We observe that the CPU power mainly affects the node power changes for both cases. Because of the dynamic power management on Theta, during each matrix loop the power adjusts dynamically,  the power increases with the increase in the matrix size from 6,000 to 20,000. The runtime mainly results in the large energy increase.
 
\begin{figure}
\center
 \includegraphics[height=1.8in, width=3.3in]{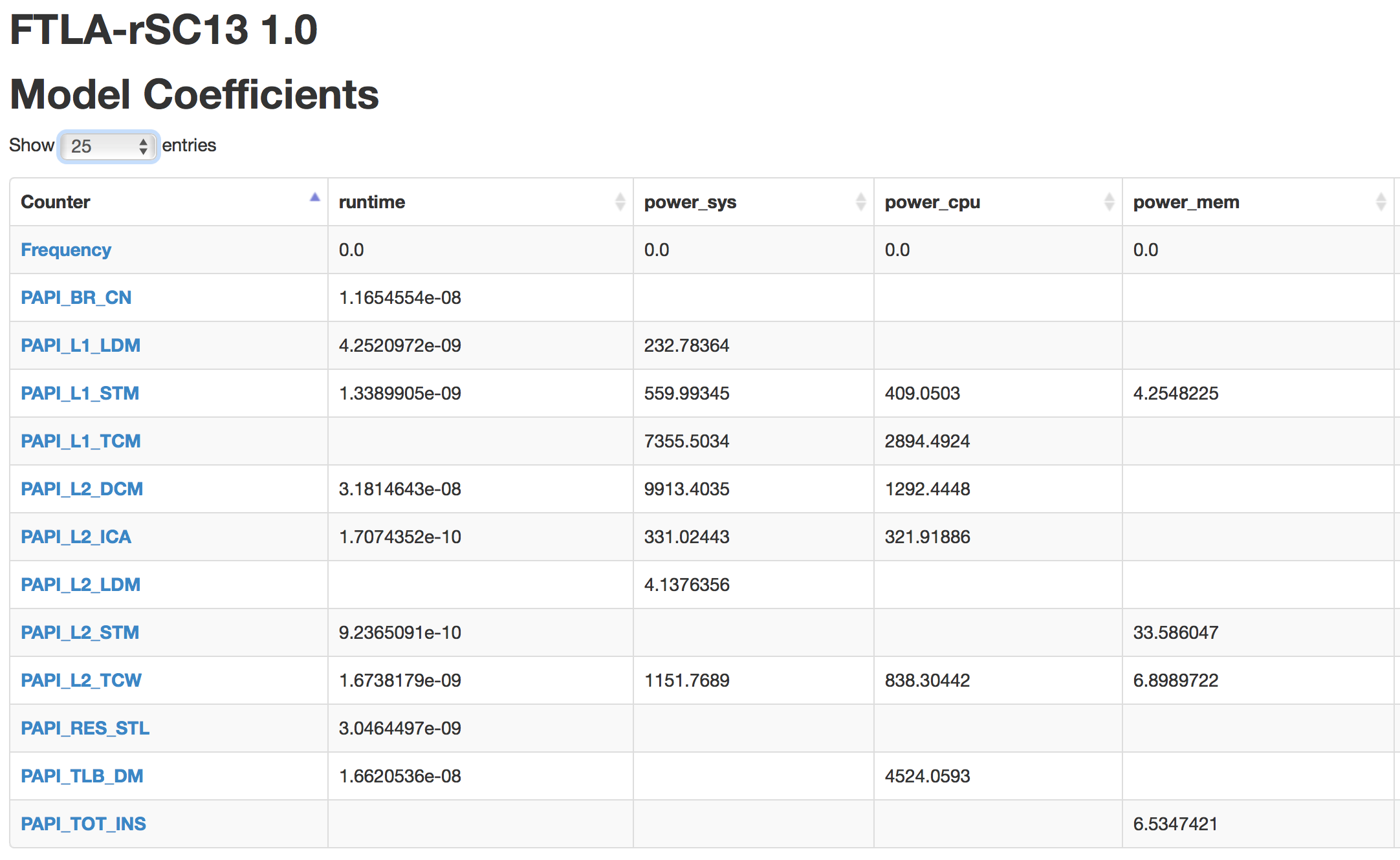}
 \caption{Four models for runtime, node power, CPU power, and memory power on Theta}
\label{fig:11}       
\end{figure}

To develop accurate models of runtime and power consumptions for the code ftla-rSC13, we use the power and performance modeling tool MuMMI from our previous work \cite{WT16} \cite{19}. We collect 40 available performance counters on Theta with different system configurations (numbers of cores: 64, 128, 256, 512, and 1024) and the number of error injections (1, 2, and 3) as a training set. We then use a Spearman correlation and principal component analysis (PCA) to identify the major performance counters ($r_1,r_2, ...,r_n (n << 40)$), which are highly correlated with the metric: runtime, system power, CPU power, or memory power. Then we use a nonnegative multivariate regression analysis to generate our four models based on the small set of major counters and CPU frequency (f), as shown in Figure \ref{fig:11}, where a numeric value is the coefficient for the counter in the corresponding model.

For the model of runtime $T$, we develop the following equation:
\begin{equation} \label{eq2}
T = \beta_0 + \beta_{1}*r_1+ \beta_{2}*r_{2}.......\beta_{n}*r_n+\beta * \frac{1}{f}.
\end{equation}

In this equation, $T$ is the component predictor used to represent the value for runtime.  The intercept is $\beta_{0}$; each $\beta_{n}$ represents 
the regression coefficient for performance counter $r_n$, and $\beta$ represents the coefficient for the CPU frequency. Equation \ref{eq2} can be used to predict the runtime for the larger numbers of error injections (4 or 5 error injections). 

Similarly, we can model CPU power consumption $P$ using the following equation:
\begin{equation} \label{eq3}
P = \alpha_0 + \alpha_{1}*r_1+ \alpha_{2}*r_{2}.......\alpha_{n}*r_n+\alpha * f^3 .
\end{equation}

In this equation, $P$ is the component predictor used to represent the value for the CPU power. 
The intercept is $\alpha_{0}$; and each $\alpha_{n}$ represents the regression coefficient for performance counter $r_n$, 
and $\alpha$ represents the coefficient for the CPU frequency. Equation \ref{eq3} can be used to predict the CPU power on larger numbers of error injections.
Similarly, a multivariate linear regression model is constructed for each metric (node power, memory power) of the same application. 
 
Table \ref{tab:4} shows the prediction error rates for the runtime and power of the application with 4 and 5 error injections using Equations \ref{eq2} and \ref{eq3}. Overall, the prediction error rates (absolute values) are less than 3.3\% in runtime. This indicates our counter-based performance models are very accurate. The prediction error rates are less than 8.9\% in node power, less than 6.3\% in CPU power, and less than 7.9\% in memory power. These performance and power models are generated from different system configurations and problem sizes, thus providing a broader understanding of the application's usage of the underlying architectures. This in turn results in more knowledge about the application's energy consumption on the given architecture.

\begin{table}
\center
\caption{Prediction error rates on Theta}
\begin{tabular}{c}
  \includegraphics[height=1in, width=3.3in]{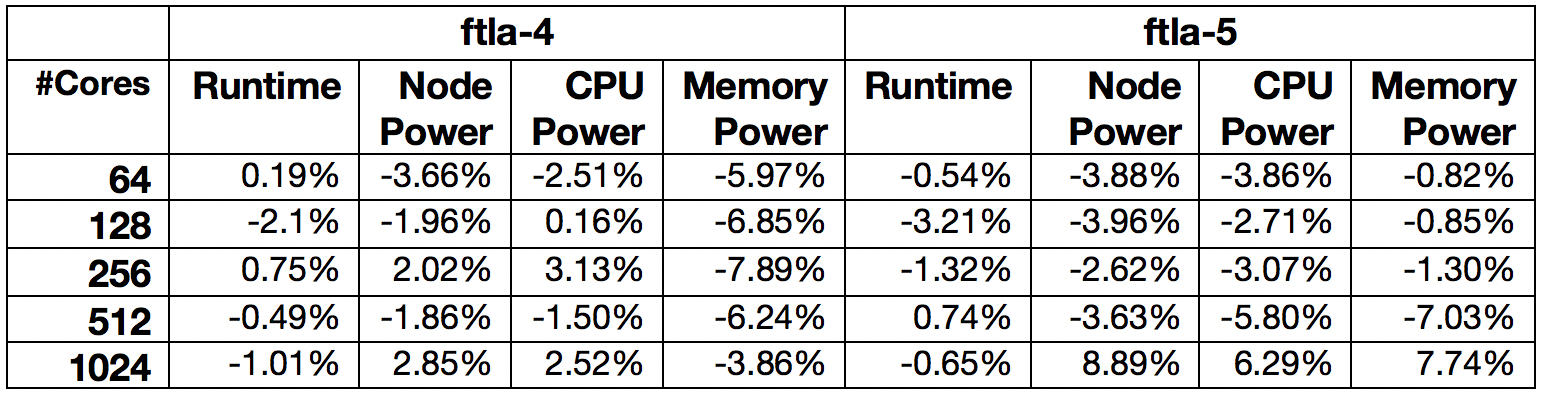}
  \end{tabular}
\label{tab:4}       
\end{table}  

\begin{figure}
\center
 \includegraphics[height=1.8in, width=3in]{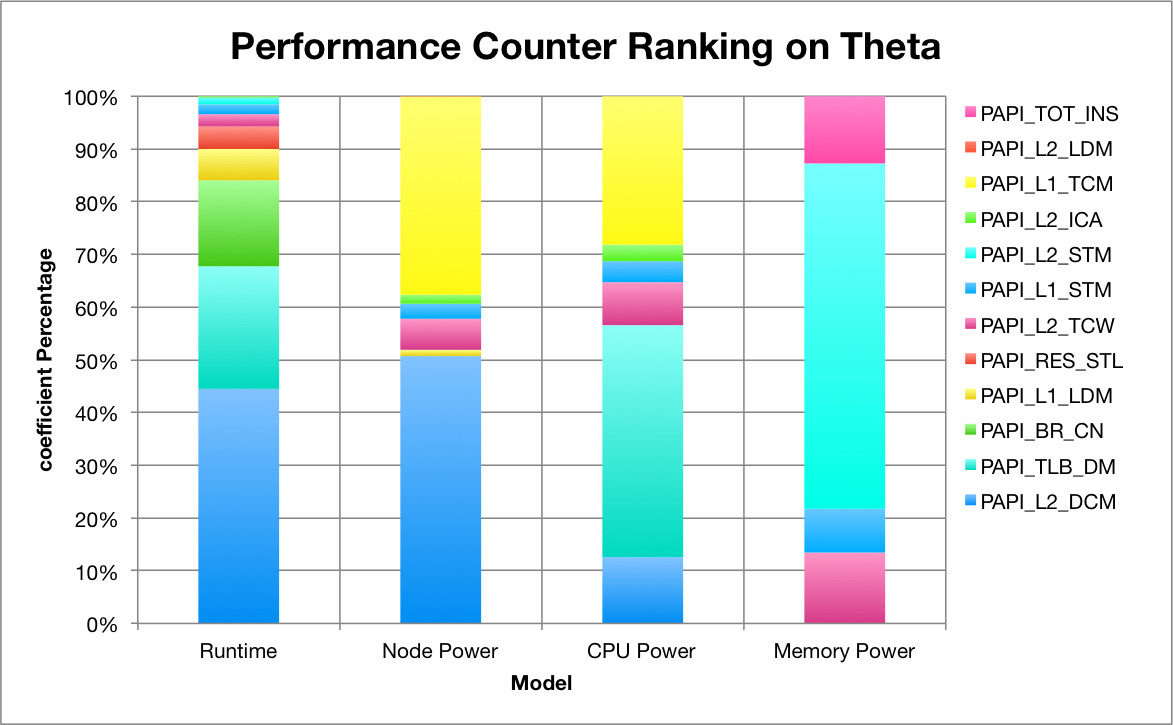}
 \caption{Counter Ranking on Theta}
\label{fig:13}       
\end{figure}

\begin{figure}
\center
 \includegraphics[height=1.2in, width=3in]{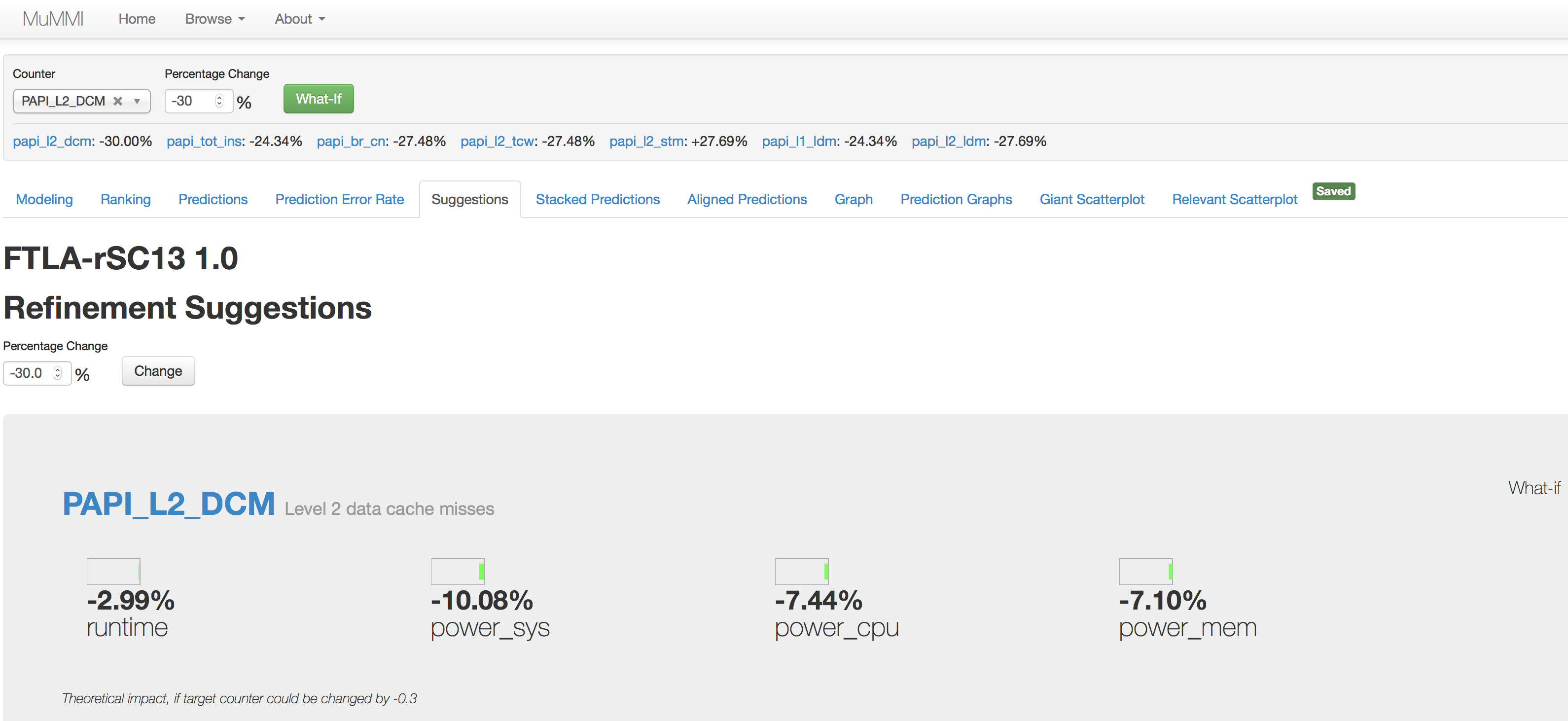}
 \caption{Theoretical prediction on Theta}
\label{fig:411}       
\end{figure}

Based on the models for runtime, node power, CPU power, and memory power, we identify the most significant performance counters for the application. Figure \ref{fig:13} shows the performance counter rankings of the four models using 12 different counters. We found that the L2\_DCM (Level 2 data cache misses) and  TLB\_DM (Data translation lookaside buffer misses) contribute most in the runtime model; L2\_DCM and L1\_TCM (Level 1 cache misses) contribute most in the node power; TLB\_DM and L1\_TCM contributes most in CPU power models; and L2\_STM (Level 2 store misses) contributes most in the memory power model. TLB\_DM is correlated with L1\_TCM. Therefore, the optimization efforts for the code should focus on the units associated with L2\_DCM, TLB\_DM, and L2\_STM on Theta. For instance, as shown in Figure \ref{fig:411}, we use our what-if prediction system based on the four model equations to predict the theoretical outcomes of the possible optimization by reducing L2\_DCM by 30\%,  the other counters may be changed based on the correlation with this counter. The theoretical improvement percentage is 2.99\% in runtime, 10.08\% in node power, 7.44\% in CPU power, and 7.10\% in memory power.  The default page size on Theta is 4 KB; but Theta supports several huge page sizes ranging from  2 MB to 2GB. In order to reduce the TLB miss (TLB\_DM), the main kernel address space is mapped with huge pages---a single 2 MB huge page requires only a single TLB entry, while the same memory, in 4 KB pages, would need 512 TLB entries. Using the huge pages will result in the application performance improvement. 

\subsection{IBM Blue Gene/Q Mira}

\begin{figure}
\center
 \includegraphics[height=1.6in, width=3.3in]{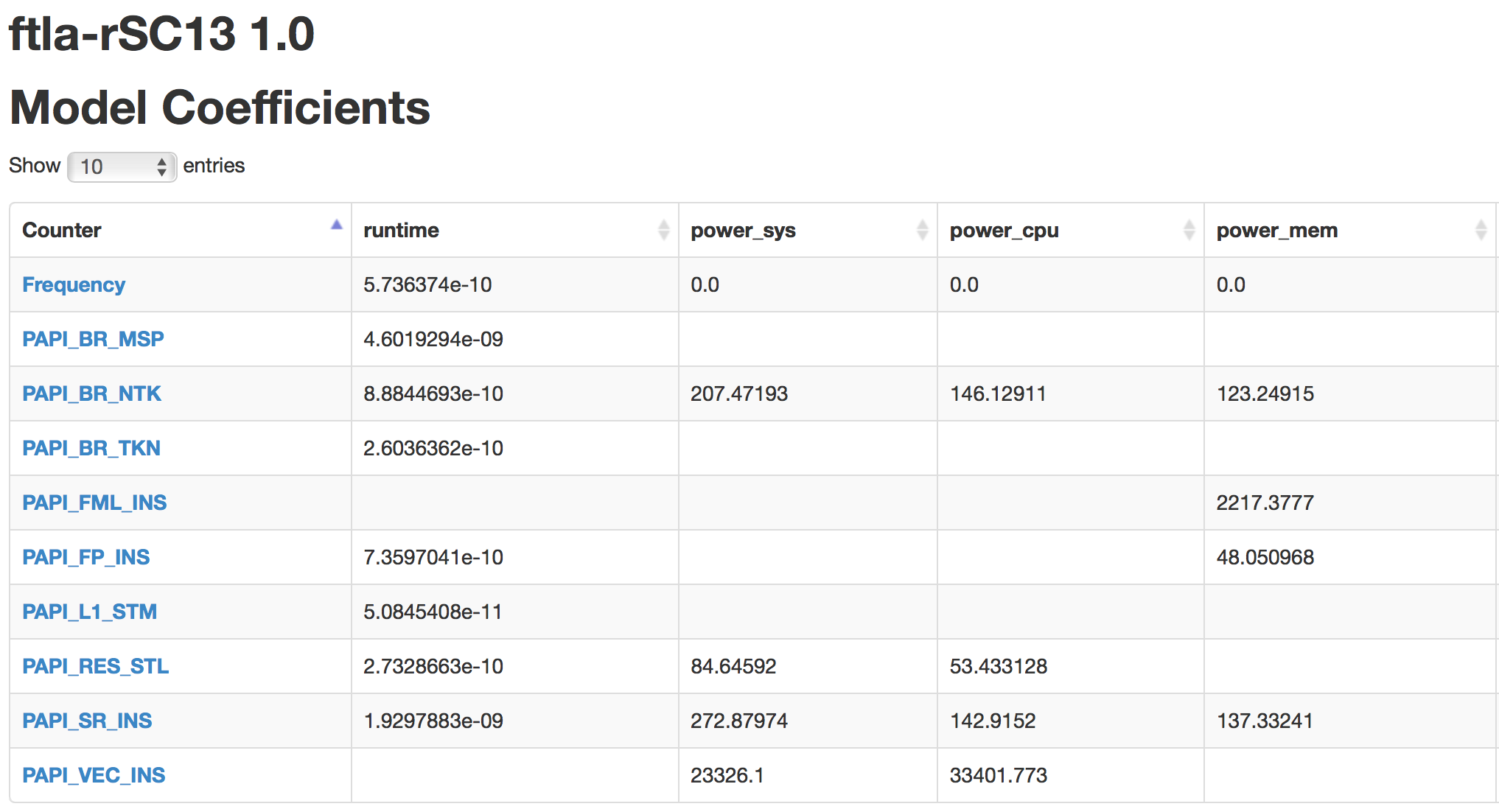}
 \caption{Four models for runtime, node power, CPU power, and memory power on Mira}
\label{fig:6}       
\end{figure}

To develop accurate models for runtime and power consumptions on Mira, we collect 40 available performance counters with different system configurations (numbers of cores: 512, 1,024, 2,048, 4,096, 8,192, and 16,384) and the number of error injections (1, 2, and 3) as a training set. We then use MuMMI to generate our four models based on the small set of major counters and CPU frequency (f), as shown in Figure \ref{fig:6}, where a numeric value is the coefficient for the counter in the corresponding model.

Table \ref{tab:2} shows the prediction error rates for the runtime and power of the application with 4 and 5 error injections using Equations \ref{eq2} and \ref{eq3}. Overall, the prediction error rates in runtime are less than 0.1\%. This indicates that our counter-based performance models are accurate. The prediction error rates are less than 5\% in node power and less than 8.7\% in CPU power; and the error rates are less than 10\% in memory power for most cases except 15.30\% for ftla-4 on 2,048 cores. 

\begin{table}
\center
\caption{Prediction error rates on Mira}
\begin{tabular}{c}
  \includegraphics[height=1in, width=3.3in]{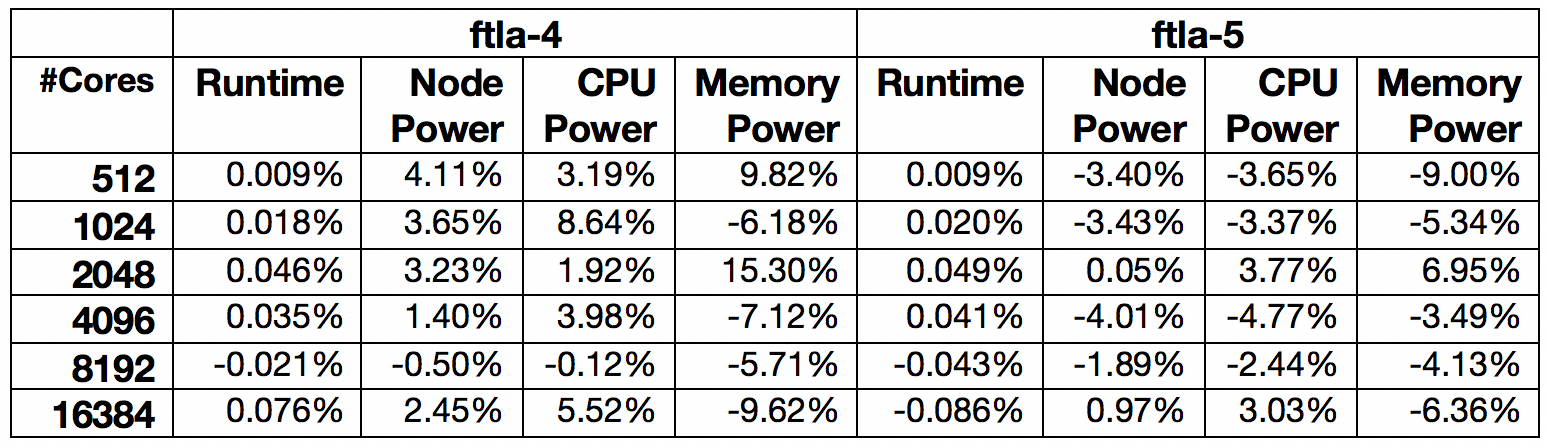}
  \end{tabular}
\label{tab:2}       
\end{table}  

\begin{figure}
\center
 \includegraphics[height=1.8in, width=3in]{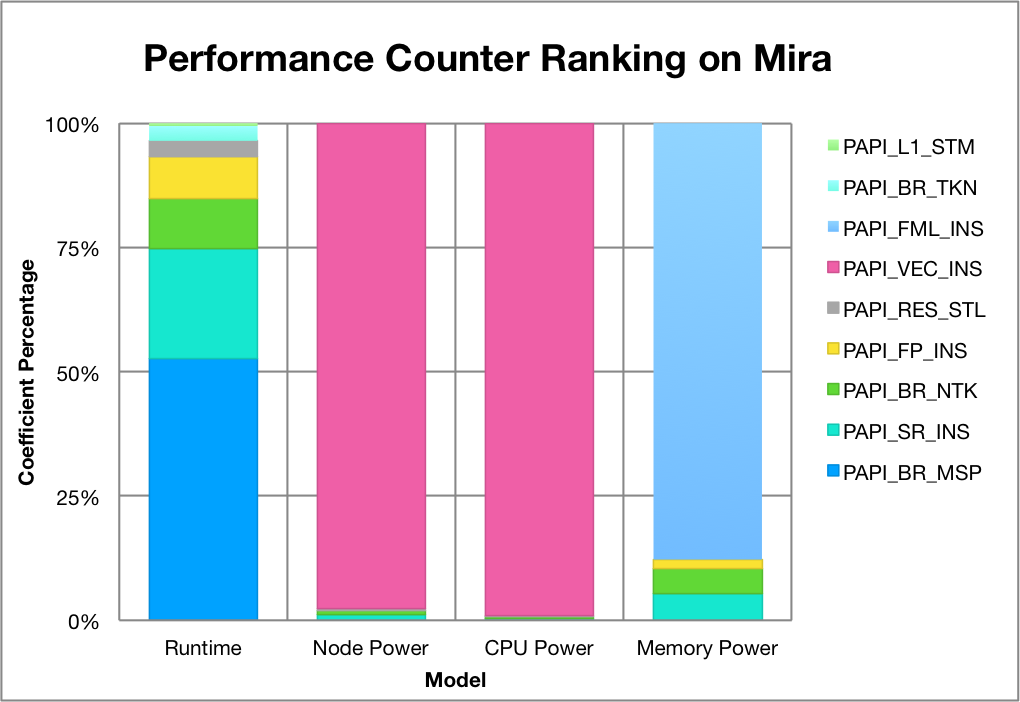}
 \caption{Counter Ranking on Mira}
\label{fig:8}       
\end{figure}

\begin{figure}
\center
 \includegraphics[height=1.2in, width=3in]{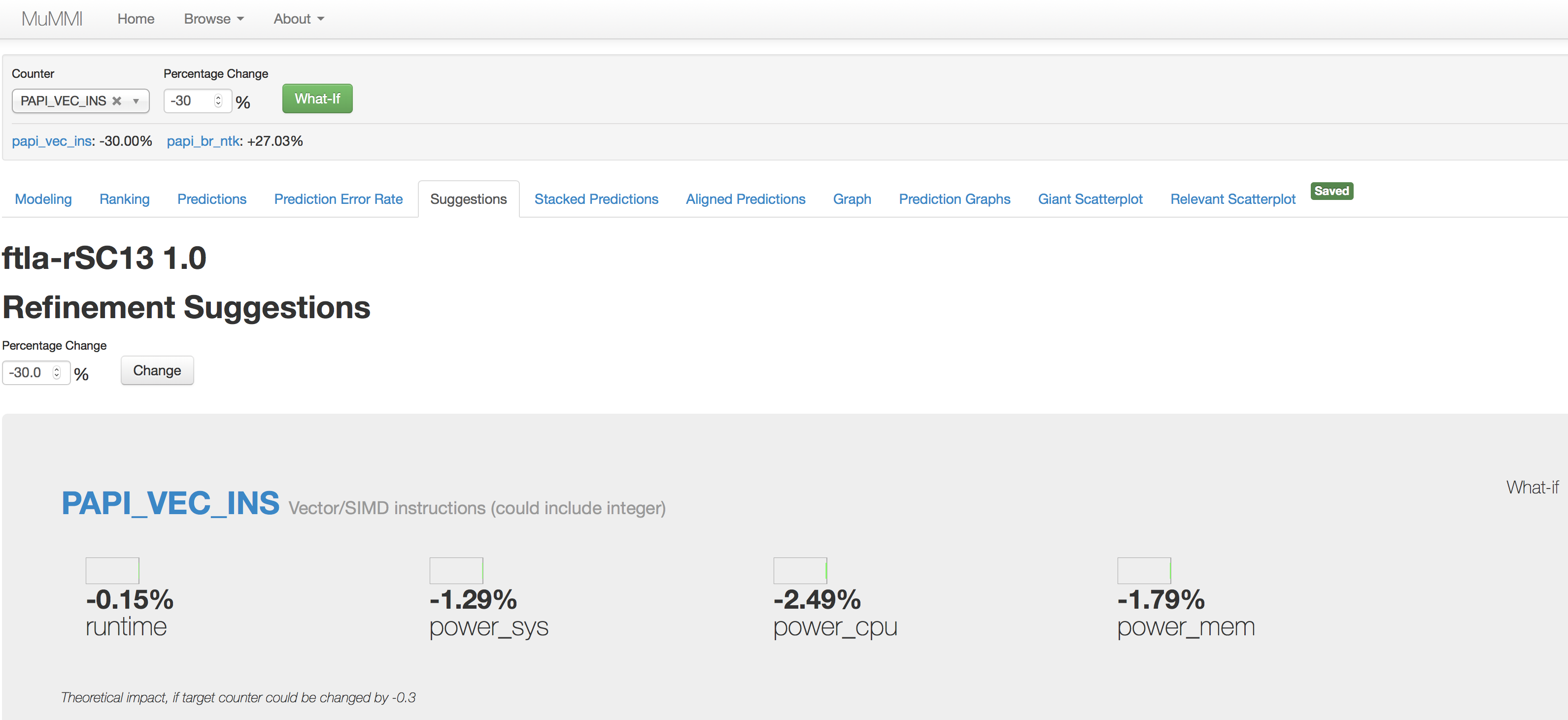}
 \caption{Theoretical prediction on Mira}
\label{fig:40}       
\end{figure}

Based on the models for runtime, node power, CPU power, and memory power, we identify the most significant performance counters for the application. Figure \ref{fig:8} shows the performance counter rankings of the four models using 9 different counters. We find that the BR\_MSP (conditional branch instructions mispredicted) contributes most in the runtime model and is correlated with the counters SR\_INS (Store instructions), BR\_TKN (Conditional branch instructions taken), FP\_INS (floating-point instructions), and RES\_STL (Cycles stalled on any resource); VEC\_INS (Vector/SIMD instructions (could include integer)) contributes most in the node power and CPU power models; and FML\_INS (floating-point multiply instructions) contributes most in the memory power model. VEC\_INS and FML\_INS are not correlated with any other counters. Therefore, the optimization efforts for the code should focus on the units associated with BR\_MSP, VEC\_INS, and FML\_INS on Mira. For instance, Mira features a quad floating-point unit  that can be used to execute four-wide SIMD instructions or two-wide complex arithmetic SIMD instructions. In order to take advantage of vector instructions supported by BG/Q processors, the compiler options -qarch-qp and -qsimd=auto may be applied to compile the code to improve the energy efficiency. For instance, as shown in Figure \ref{fig:40}, we use our what-if prediction system based on the four model equations to predict the theoretical outcomes of the possible optimization. By accelerating VEC\_INS by 30\%, the theoretical improvement percentage is 0.15\% in runtime, 1.29\% in node power, 2.49\% in CPU power, and 1.79\% in memory power.

\subsection{Intel Haswell Shepard}

\begin{figure}
\center
 \includegraphics[height=1.8in, width=3.3in]{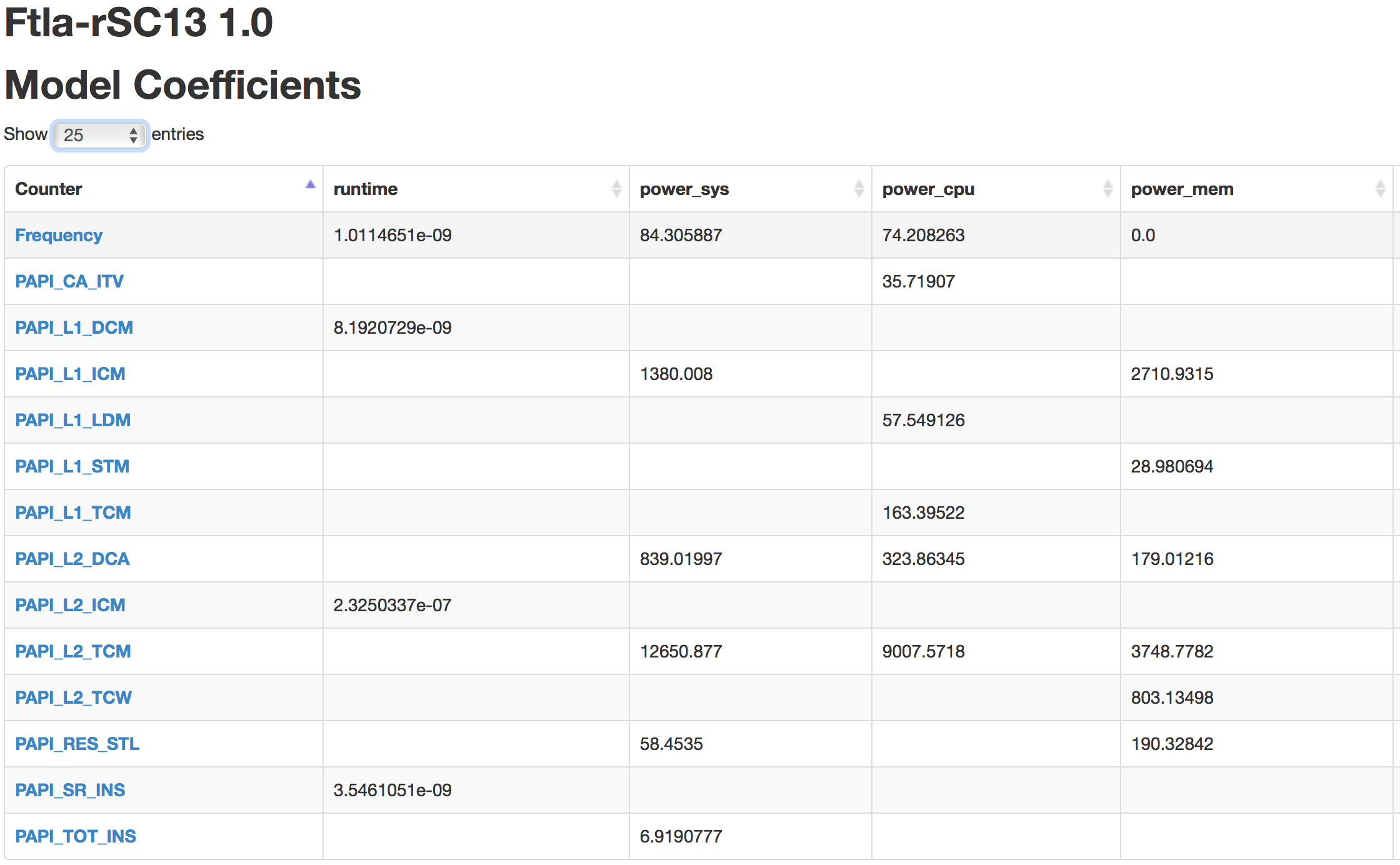}
 \caption{Four models for Runtime, node power, CPU power, and memory power on Shepard}
\label{fig:20}       
\end{figure}

To develop accurate models for runtime and power consumptions on Shepard, we use MuMMI to generate the four models based on the small set of major counters and CPU frequency (f), as shown in Figure \ref{fig:20} with the training dataset for different system configurations (numbers of cores: 32, 64, 128, 256, 512, and 1024) and the number of error injections (1, 2, and 3).

Table \ref{tab:5} shows the prediction error rates for the runtime and power of the application with 4 and 5 error injections using Equations \ref{eq2} and  \ref{eq3}. Overall, the prediction error rates (absolute values) are less than 0.25\% for runtime. These results indicate that our counter-based performance models are accurate.  The prediction error rates are less than 7.3\% in node power, and less than 6.6\% in CPU power; and the prediction error rates in memory power are less than 7.46\% for most cases except 16.57\% for ftla-4 and 12.88\% for ftla-5 on 256 cores.

\begin{table}
\center
\caption{Prediction error rates on Shepard}
\begin{tabular}{c}
  \includegraphics[height=1in, width=3.3in]{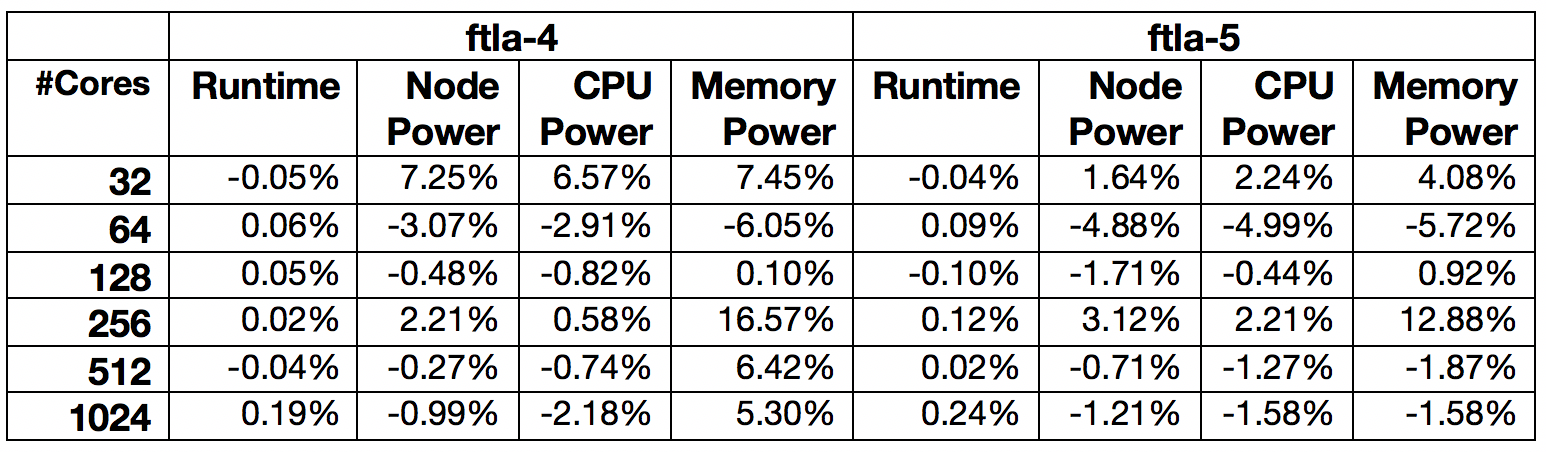}
  \end{tabular}
\label{tab:5}       
\end{table}  
 
\begin{figure}
\center
 \includegraphics[height=1.8in, width=3in]{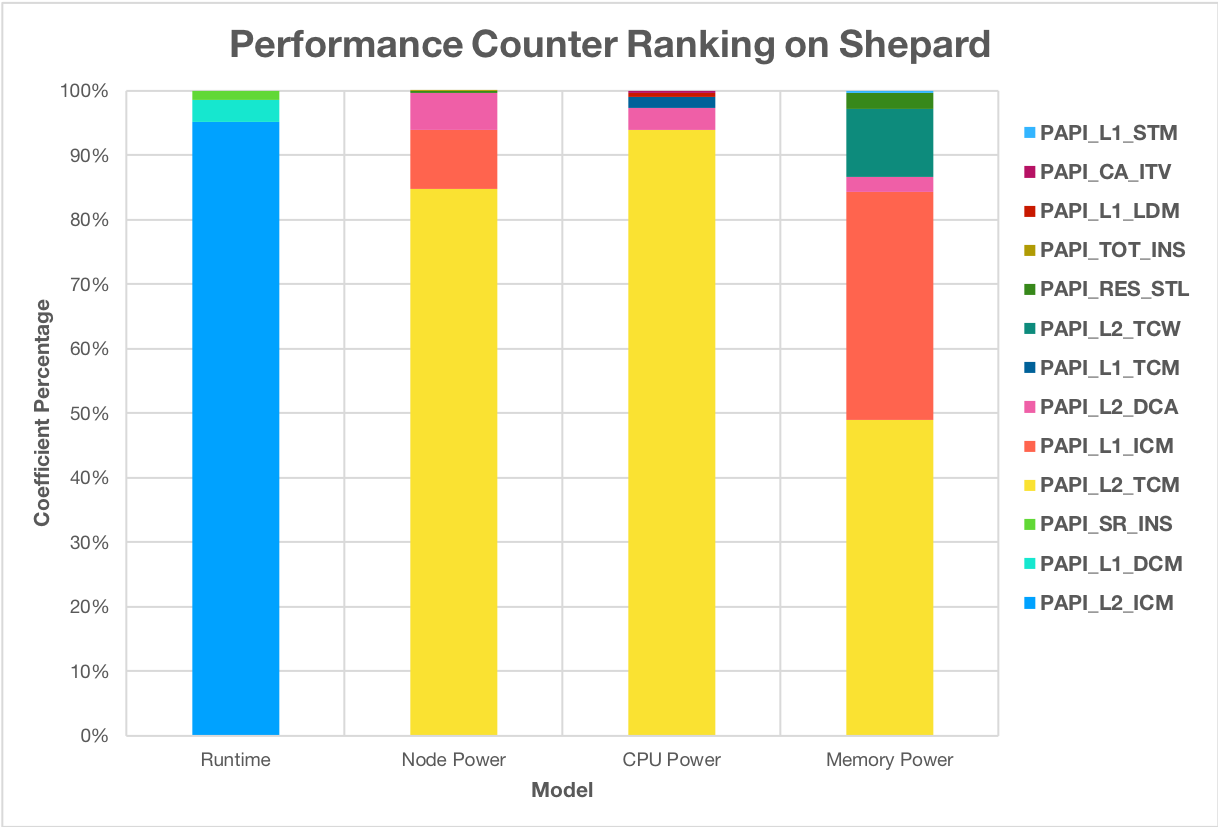}
 \caption{Counter ranking on Shepard}
\label{fig:23}       
\end{figure}

\begin{figure}
\center
 \includegraphics[height=1.2in, width=3in]{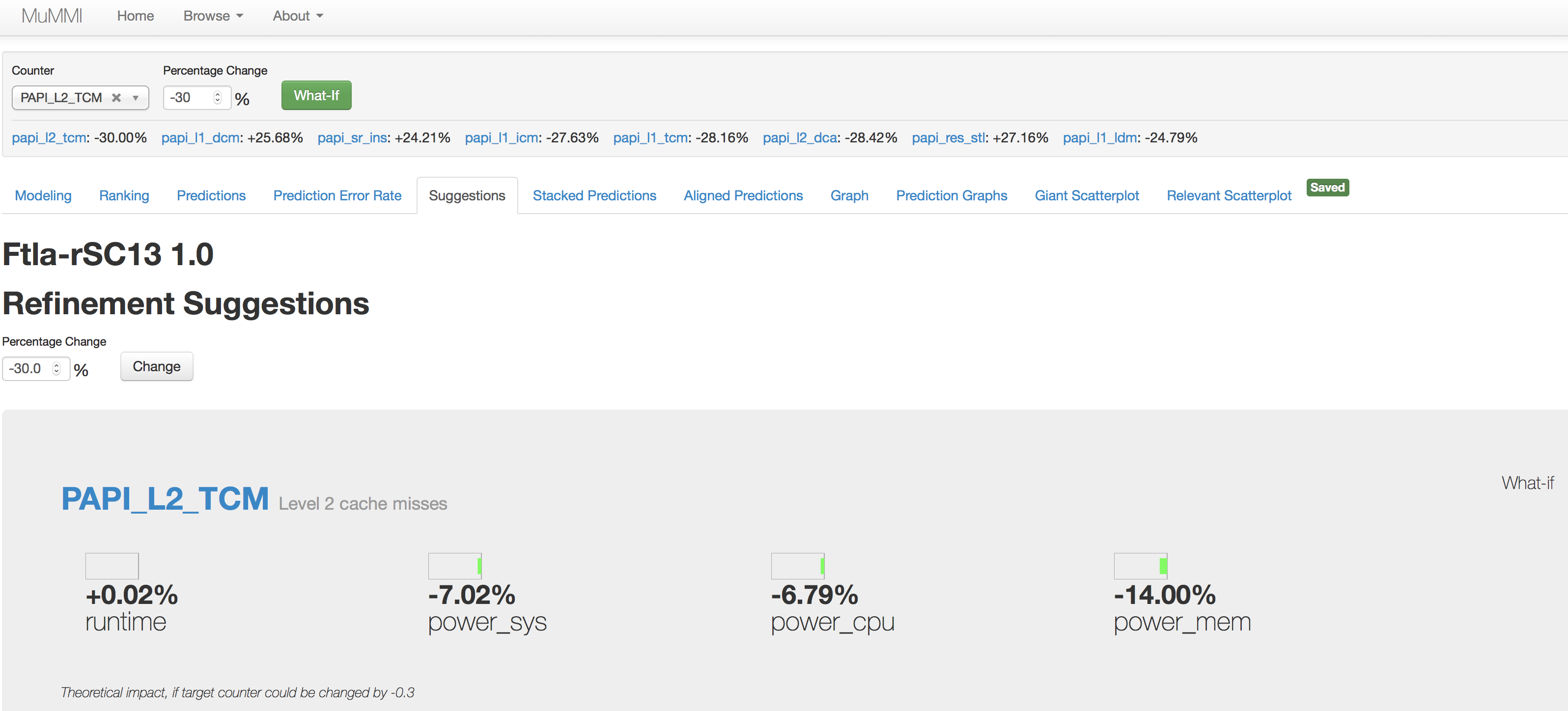}
 \caption{Theoretical prediction on Shepard}
\label{fig:92}       
\end{figure}

Based on the models for runtime, node power, CPU power, and memory power, we identify the most significant performance counters for the application. Figure \ref{fig:23} shows the performance counter rankings of the four models using 13 different counters. We found that the L2\_ICM (Level 2 instruction cache misses) and  L1\_DCM (Level 1 data cache misses) contribute most in the runtime model; L2\_TCM (Level 2 cache misses) and L1\_ICM (Level 1 instruction cache misses) contribute most in the node power; L2\_TCM and L1\_TCM contributes most in CPU power models; and L2\_TCM  and L1\_ICM contribute most in memory power model. L2\_ICM is correlated with L1\_TCM, and L2\_TCM is correlated with L1\_ICM. Therefore, the optimization efforts for the code should focus on the units associated with L2 and L1 caches on Shepard. For instance, as shown in Figure \ref{fig:92}, we use our what-if prediction system based on the four model equations to predict the theoretical outcomes of the possible optimization by reducing L2\_TCM by 30\%,  the other counters may be changed based on the correlation with this counter. The theoretical improvement percentage is -0.02\% in runtime, 7.02\% in node power, 6.79\% in CPU power, and 14\% in memory power. For instance, loop optimization methods such as loop blocking and unrolling may help improve the cache locality. 

\section{Modeling and Prediction Using 10 Machine Learning Methods}

In this section, we use 10 machine learning (ML) methods from the R caret package\cite{CAR} \cite{KJ13} to model and predict performance and power of FTLA and HDC. Our methodology is as follows. First, we use the datasets for FTLA or HDC as input to split the data into the training and test datasets based on the 80/20\% rule, and find out what the training and test datasets are by setting the seed 3456 of R's random number generator set.seed() so that creating the random objects can be reproduced. Second, we apply the same training and test datasets to the 10 ML methods. Third, we use the same training and test datasets to build the performance and power models using MuMMI online.  Finally, we compare the prediction error rates for these methods using violin plot from R violin package \cite{VIO} which is a combination of a box plot and a kernel density plot to visualize the distribution of the prediction error rates. 

The 10 ML methods are Random Forest (RF), Gaussian Process with Radial Basis Function (GP), eXtreme Gradient Boosting (xGB), Stochastic gradient boosting (Sgb), Cubist (Cub), Ridge Regression (RR),  k-Nearest Neighbors (kNN), Support Vector Machines with Linear Kernel (SVM), Conditional Inference Tree (CIT), and Multivariate Adaptive Regression Spline (MAR).

RF is short for random forests \cite{LW18} for regression or classification based on a forest of trees using random inputs, which was constructed in \cite{BL01} as a tree-based model. A random forest model achieves the variance reduction by selecting strong, complex learners that exhibit low bias. This ensemble of many independent, strong learners yields an improvement in error rates.

GP is short for Gaussian Process \cite{WR95} with Radial Basis Function \cite{KSH}, which is based on the prior assumption that adjacent observations should convey information about each other. It is assumed that the observed variables are normal, and that the coupling between them takes place by means of the covariance matrix of a normal distribution. Using the kernel matrix as the covariance matrix is a convenient way of extending Bayesian modeling of linear estimators to nonlinear situations. 

xGB is short for the eXtreme gradient boosting\cite{CH19}, which is an efficient implementation of the gradient boosting framework in \cite{CG16}. It provides a sparsity aware algorithm for handling sparse data and a theoretically justified weighted quantile sketch for approximate learning.

Sgb is short for Stochastic gradient boosting \cite{GB19}, which is an implementation of extensions to AdaBoost algorithm \cite{FS96} and gradient boosting machine \cite{FR01}. It includes regression methods for least squares, absolute loss, t-distribution loss, quantile regression, logistic, multinomial logistic, Poisson, Cox proportional hazards partial likelihood, AdaBoost exponential loss, Huberized hinge loss, and Learning to Rank measures.
 
Cub is short for cubist \cite{KW20} which is a regression modeling using rules with added instance-based corrections that combines the ideas in \cite{QR92} and \cite{QR93}. A cubist regression model is to fit for each rule based on the data subset defined by the rules. The set of rules are pruned or possibly combined, and the candidate variables for the linear regression models are the predictors that were used in the parts of the rule that were pruned away.

RR is short for Ridge Regression \cite{ZH18} \cite{HO70}, which adds a penalty on the sum of the squared regression parameters to create biased regression models. It reduces the impact of collinearity on model parameters. Combatting collinearity by using biased models may result in regression models where the overall mean squared error is competitive.

kNN is short for k-Nearest Neighbors \cite{KJ13}, which imply predicts a new sample using the k-closest samples from the training set. To predict a new sample for regression, It identifies that sample's k-Nearest Neighbors in the predictor space. The predicted response for the new sample is then the mean of the k neighbors' responses.  

SVM is short for Support Vector Machines with Linear Kernel \cite{KSH}, which is the kernlab's implementation of support vector machines \cite{VV98}. It chooses a linear function in the feature space by optimizing some criterion over the sample.

CIT is short for the conditional inference tree \cite{HH20}, which embeds tree-structured regression models into a well defined theory of conditional inference procedures. This non-parametric class of regression trees is applicable to all kinds of regression problems, including nominal, ordinal, numeric, censored as well as multivariate response variables and arbitrary measurement scales of the covariates.

MAR is short for Multivariate Adaptive Regression Splines\cite{MS19}, which builds a regression model using the techniques in \cite{FJ91}. It is a form of regression analysis that is an extension to linear regression that captures nonlinearities and interactions between variables.

\subsection{FTLA}
For FTLA with the fixed problem size (matrix sizes from 6,000 to 20,000 with a stride of 2000 and a block size of 100, strong scaling), we ran the FTLA with five numbers of error injections (1, 2, 3, 4, and 5) on six different numbers of cores (32, 64, 128, 256, 512, and 1024) with five CPU frequency settings (1.2, 1.5, 1.8, 2.1 and 2.3 GHz) to collect the total 144 data samples on Shepard. Each data sample includes 53 variables such as application name, system name, number of cores, matrix sizes, stride size, block size, number of error injections, CPU frequency, 32 available performance counters, runtime, system power, CPU power, memory power, and so on. The 32 performance counters are TOT\_CYC, TOT\_INS, L1\_TCM, L2\_TCM, L3\_TCM, CA\_SHR, BR\_CN, BR\_TKN, BR\_NTK, BR\_MSP, CA\_CLN, CA\_ITV, RES\_STL, L2\_TCA, L1\_STM, L2\_TCW, L1\_LDM, L2\_DCA, L2\_DCR, L2\_DCW, L1\_ICM, BR\_INS, L1\_DCM, L2\_ICA, TLB\_DM, TLB\_IM, L2\_DCM, L2\_ICM, LD\_INS, SR\_INS, L2\_LDM, L2\_STM, then TOT\_CYC is used to normalize all the performance counters. The metrics for performance and power are runtime, node power, CPU power and memory power. We split the data as training and test datasets with the 80/20\% rule so that the training dataset consists of 116 samples, and the test dataset consists of 28 samples. For the fair comparison, we apply the same training and test datasets to all modeling methods.

\begin{figure}
\center
 \includegraphics[height=2in, width=3in]{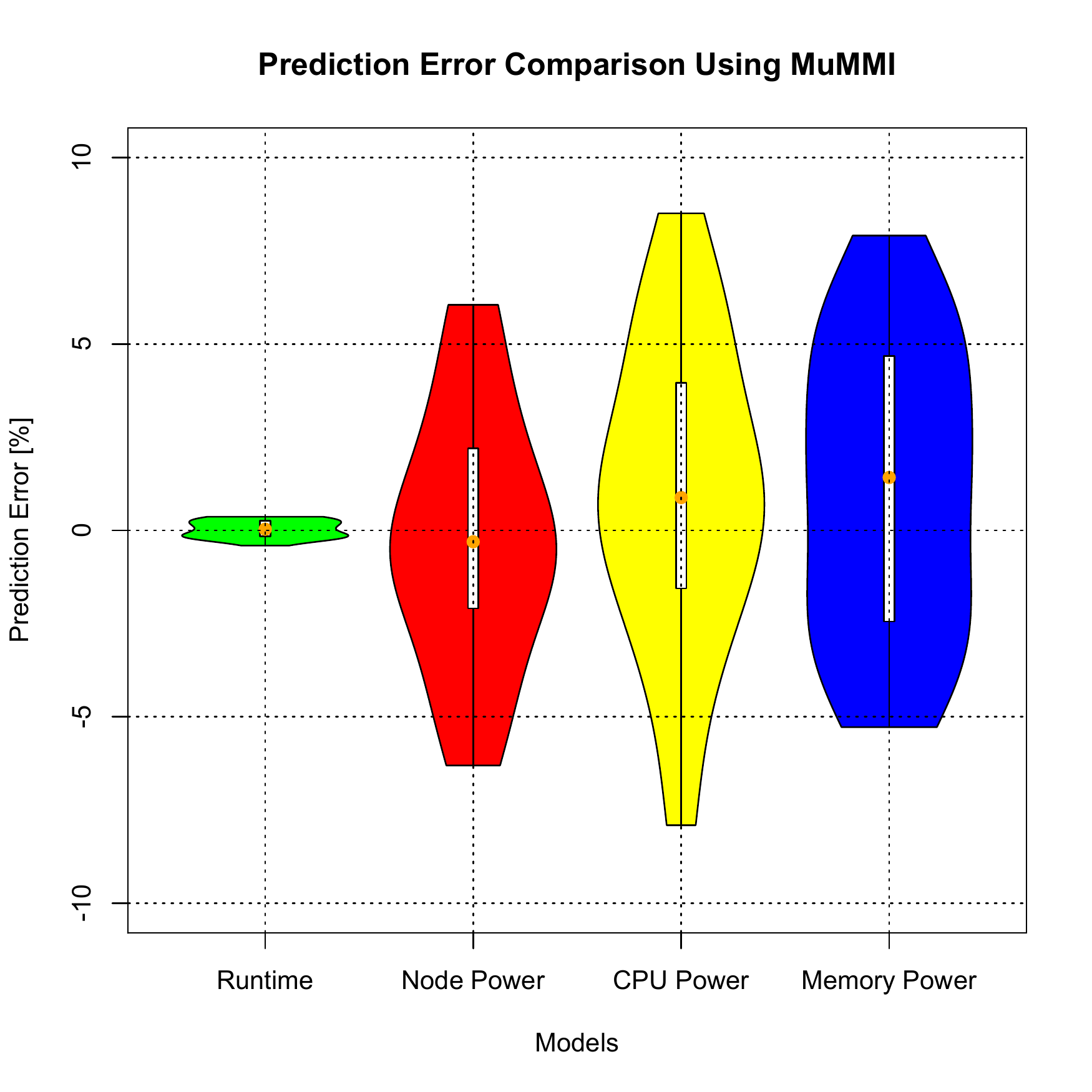}
 \caption{Prediction error rates for FTLA using MuMMI}
\label{fig:43}       
\end{figure}

Before we compare the prediction error rates for MuMMI with 10 ML methods, we look at MuMMI first. Figure ~\ref{fig:43} shows the prediction error rates for the models in runtime, node power, CPU power, and memory power using MuMMI. The prediction error rates are between -0.41\% and 0.37\% in runtime and between -6.31\% and 6.06\% in node power. The mean error rate is 0.03\% in runtime, -0.26\% in node power, 0.93\% in CPU power, and 1.00\% in memory power. Overall, this prediction is very accurate in runtime and power using MuMMI.

\begin{figure}
\center
 \includegraphics[height=2.4in, width=3.3in]{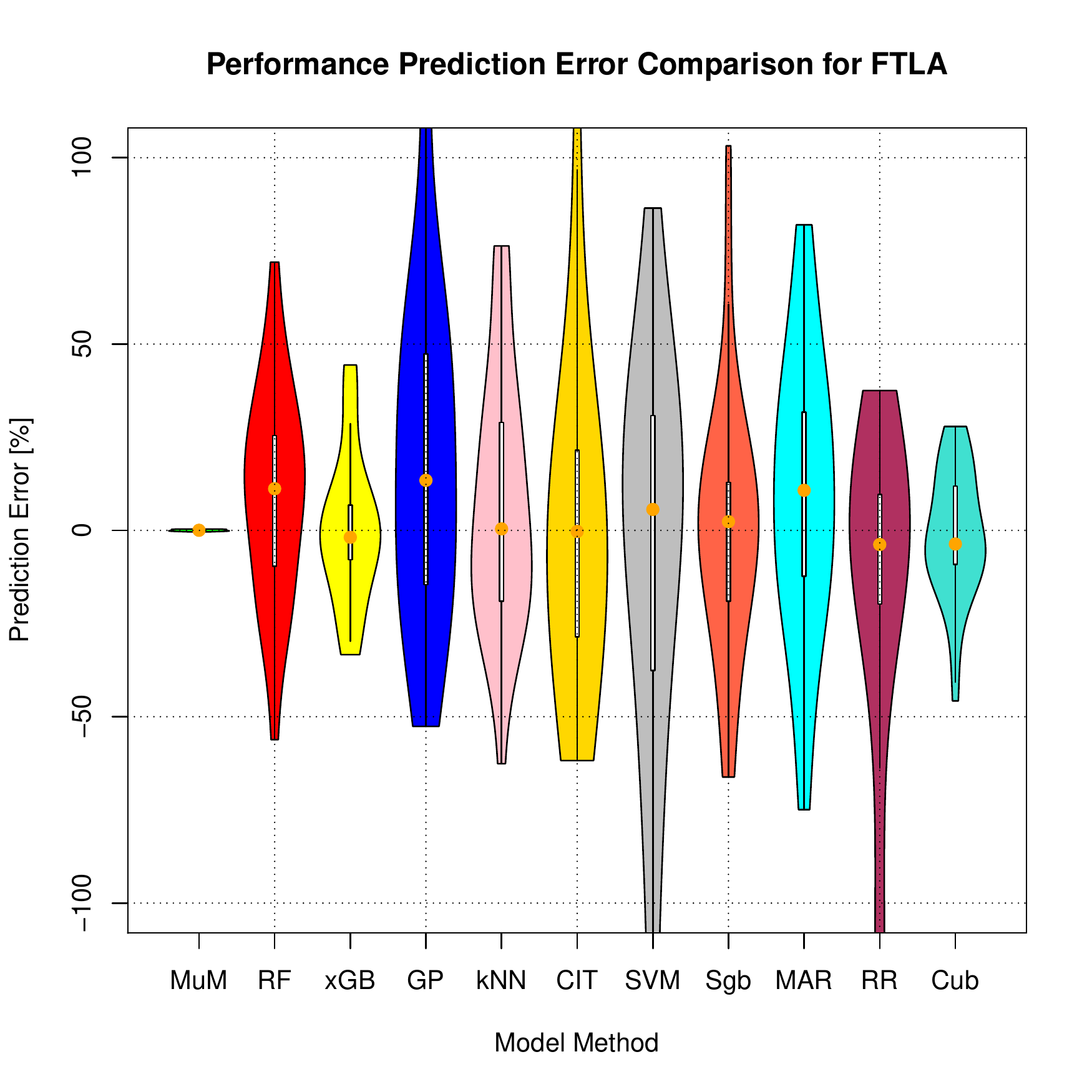}
 \caption{Prediction error rates (Runtime) for FTLA }
\label{fig:41}       
\end{figure}

For the sake of simplicity, we use ML methods to model the performance and node power only in this paper. Figure \ref{fig:41} shows the performance prediction error rates using 10 ML methods and MuMMI (MuM). These violin plots visualize the distribution of the prediction error rates for each method. Based on these error rates, we observe that Cubist (Cub) and eXtreme Gradient Boosting (xGB) resulted in the lowest error rates in performance among 10 ML methods, and for other ML methods, the maximum error rates are more than 50\%. Overall, MuMMI outperformed all of them in performance prediction.    

\begin{figure}
\center
 \includegraphics[height=2.4in, width=3.3in]{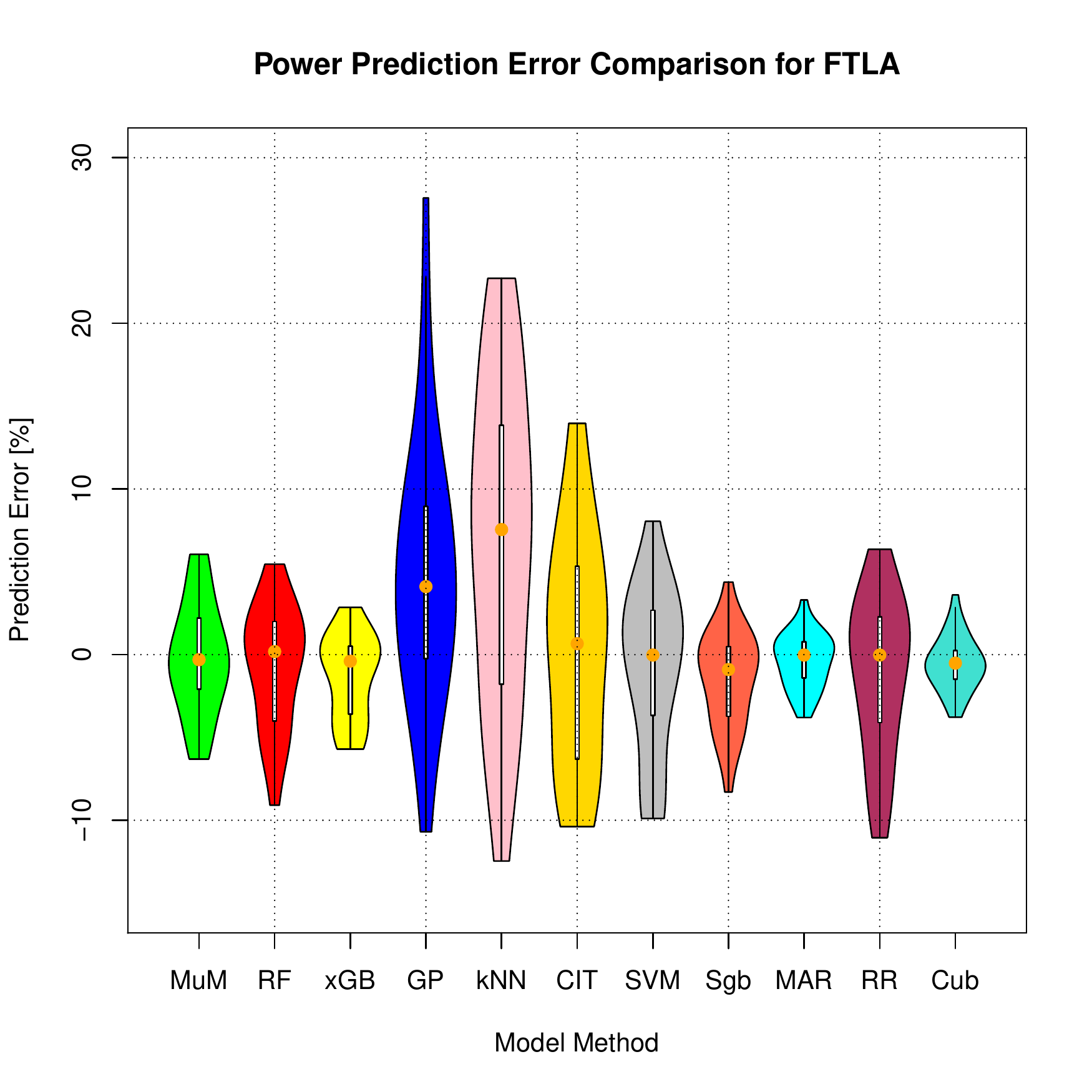}
 \caption{Prediction error rates (Node Power) for FTLA }
\label{fig:42}       
\end{figure}

Figure \ref{fig:42} shows the node power prediction error rates using MuMMI and 10 ML methods. Based on these error rates, we observe that they are between -15\% and 30\%. Multivariate Adaptive Regression Spline (MAR), Cubist (Cub) and eXtreme Gradient Boosting (xGB) resulted in the lowest error rates in node power among 10 ML methods and outperformed MuMMI although the node power prediction error rates using MuMMI are between -6.31\% and 6.06\%.

For the sake of simplicity, we choose four ML methods: Cubist, eXtreme Gradient Boosting, Random Forests, and Multivariate Adaptive Regression Spline to do an in-depth analysis in performance and power modeling and prediction.

\begin{figure}
\center
 \includegraphics[height=2in, width=3in]{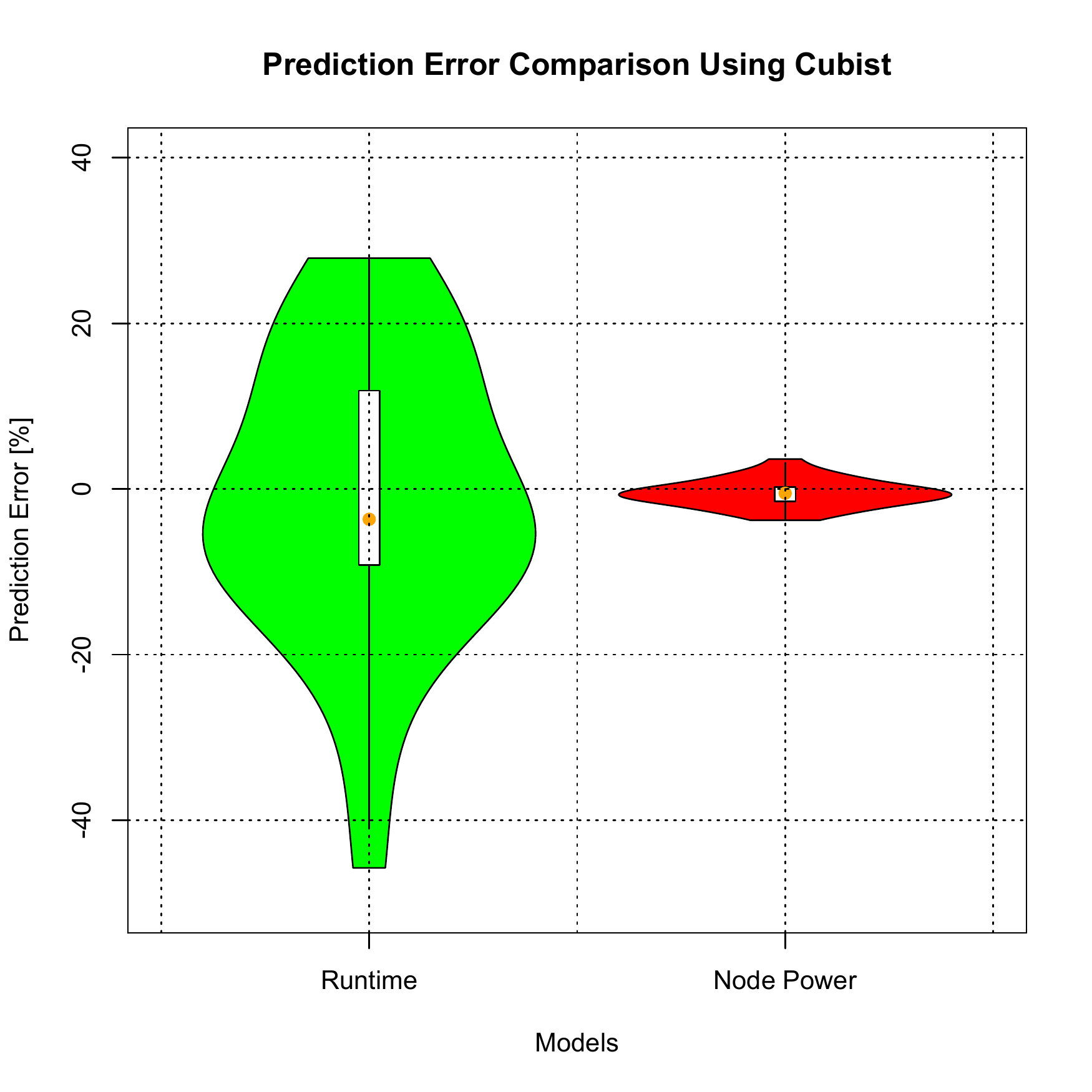}
 \caption{Prediction error rates using Cubist}
\label{fig:46}       
\end{figure}

\begin{figure}
\center
 \includegraphics[height=2in, width=3in]{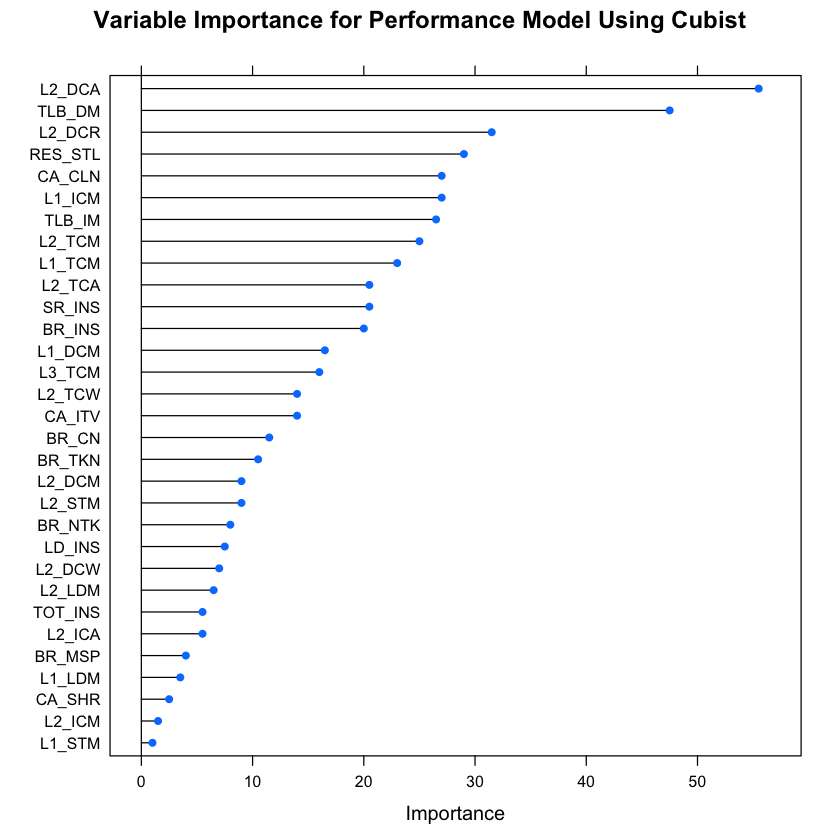}
 \caption{Variable importance for performance model of FTLA}
\label{fig:60}       
\end{figure}

\begin{figure}
\center
 \includegraphics[height=2in, width=3in]{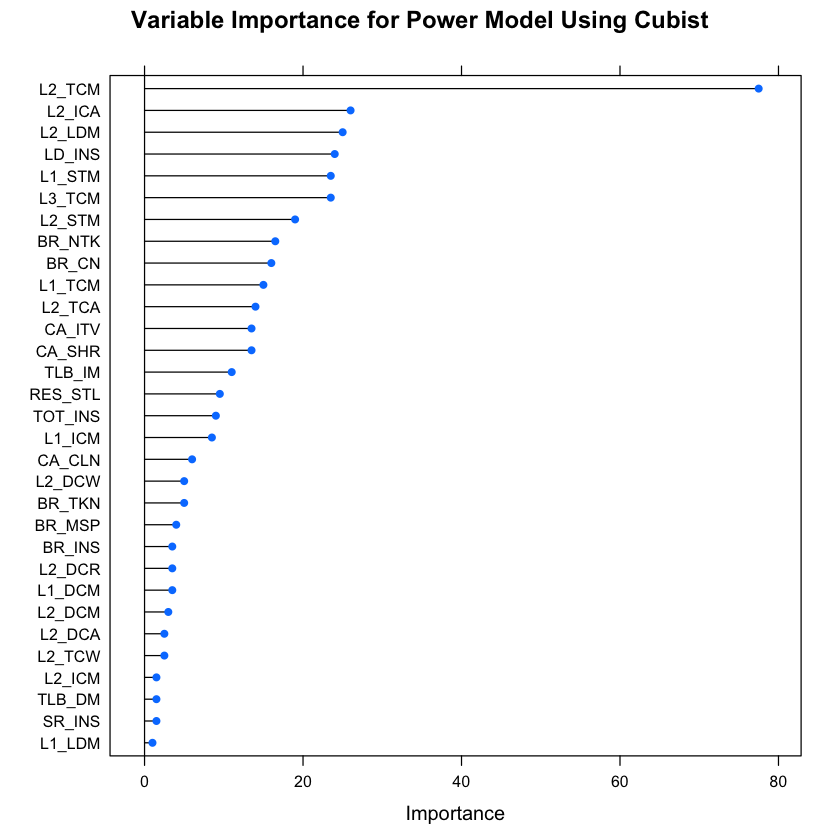}
 \caption{Variable importance for node power model of FTLA}
\label{fig:61}       
\end{figure}

Figure ~\ref{fig:46} shows the prediction error rates for runtime and node power models using Cubist. The prediction error rates are between -45.75\% and 27.88\% in runtime, and between -3.78\% and 3.61\% in node power. The mean error rate is -1.16\% in runtime and -0.54\% in node power. The performance model under-predicted for the worst case. Let's look at how the variables contribute in performance and node power models. To measure predictor importance for Cubist models \cite{KJ13}, we can enumerate how many times a predictor variable was used in either a linear model or a split and use these tabulations to get a rough idea the impact each predictor has on the model. Figure ~\ref{fig:60} shows the top 31 most important predictors for performance model of FTLA, where the x-axis is the total usage of the predictor (i.e., the number of times it was used in a split or a linear model). The larger the importance value, the more important the predictor is in relating the latent predictor structure to the response. Very small importance values are likely not considered to contain predictive information for the response and should be considered as candidates for removal from the model. Figure ~\ref{fig:61} shows the the top 31 most important predictors for node power model of FTLA. We observe that the top 3 counters in performance model are L2\_DCA, TLB\_DM, and L2\_DCR; the top 3 counters in power model are L2\_TCM, L2\_ICA, and L2\_LDM. Overall, L2 cache and TLB mainly impact the performance and power by using Cubist, however, it is interesting to observe that the top 3 counters in performance model are in the bottom of the counter list in power model, and the top 3 counters in power model are in the bottom of the counter list in performance model.

\begin{figure}
\center
 \includegraphics[height=2in, width=3in]{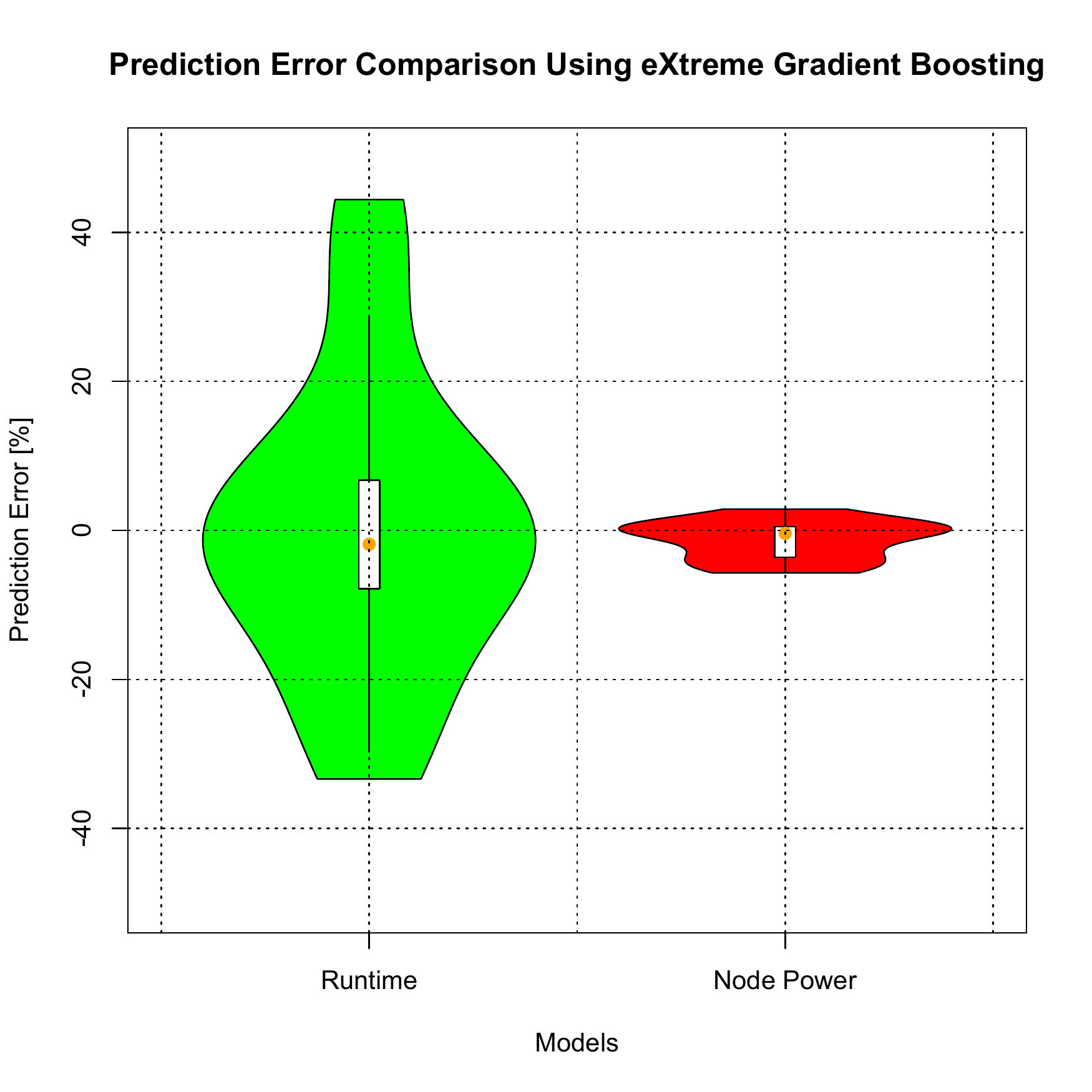}
 \caption{Prediction error rates using eXtreme Gradient Boosting}
\label{fig:45}       
\end{figure}

\begin{figure}
\center
 \includegraphics[height=2in, width=3in]{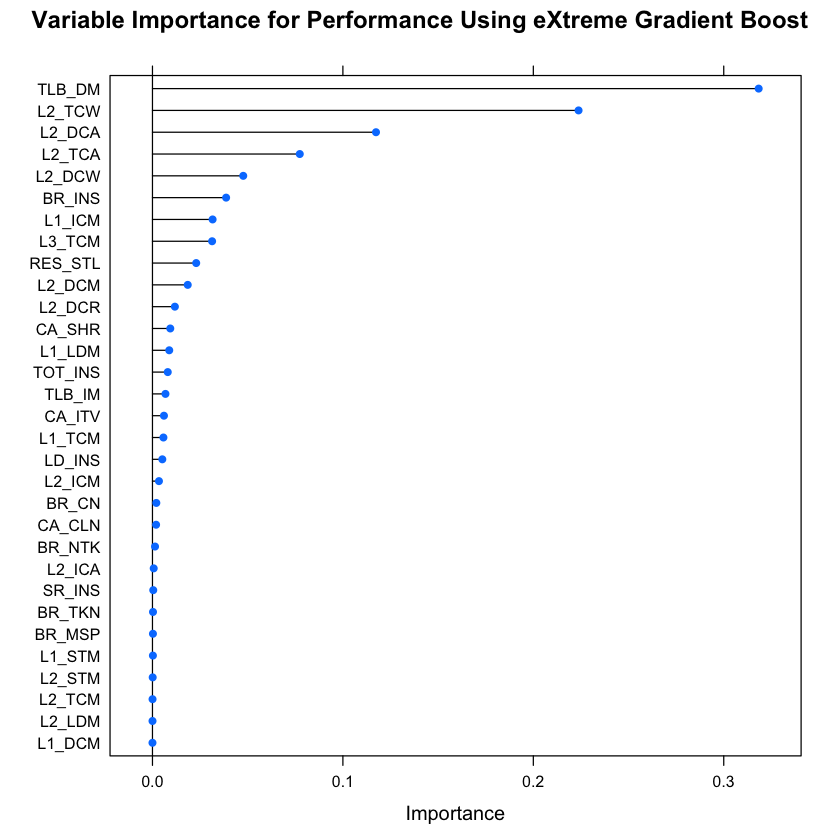}
 \caption{Variable importance for performance model of FTLA}
\label{fig:62}       
\end{figure}

\begin{figure}
\center
 \includegraphics[height=2in, width=3in]{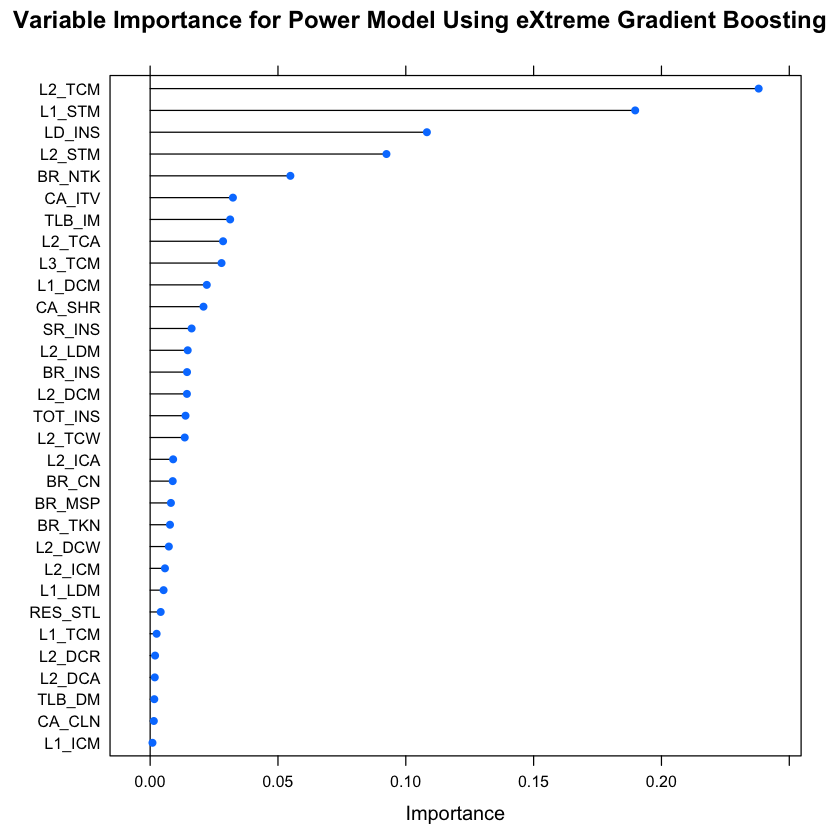}
 \caption{Variable importance for node power model of FTLA}
\label{fig:63}       
\end{figure}

Figure ~\ref{fig:45} shows the prediction error rates for runtime and node power models using eXtreme Gradient Boosting. The prediction error rates are between -33.35\% and 44.38\% in runtime, and between -5.71\% and 2.85\% in node power. The mean error rate is 0.27\% in runtime and -1.32\% in node power. Variable importance for boosting is a function of the reduction in squared error. Figure ~\ref{fig:62} shows the top 31 most important predictors for performance model of FTLA. Figure ~\ref{fig:63} shows the variable importance for node power model of FTLA. We observe that the top 3 counters in performance model are TLB\_DM, L2\_TCW, and L2\_DCA; the top 3 counters in power model are L2\_TCM, L1\_STM, and LD\_INS. Overall, TLB, L2 cache and L1 cache mainly impact the performance and power by using eXtreme Gradient Boosting. Similarly, we observe that the top 3 counters in performance model are in the bottom of the counter list in power model, and the top 3 counters in power model are in the bottom of the counter list in performance model. Contrasting the importance results to Cubist in Figure ~\ref{fig:60} and \ref{fig:61}, we see that 2 of the top 5 counters are the same (L2\_DCA and TLB\_DM in performance model; L2\_TCM and L1\_STM in power model), however, the importance orderings are much different.

\begin{figure}
\center
 \includegraphics[height=2in, width=3in]{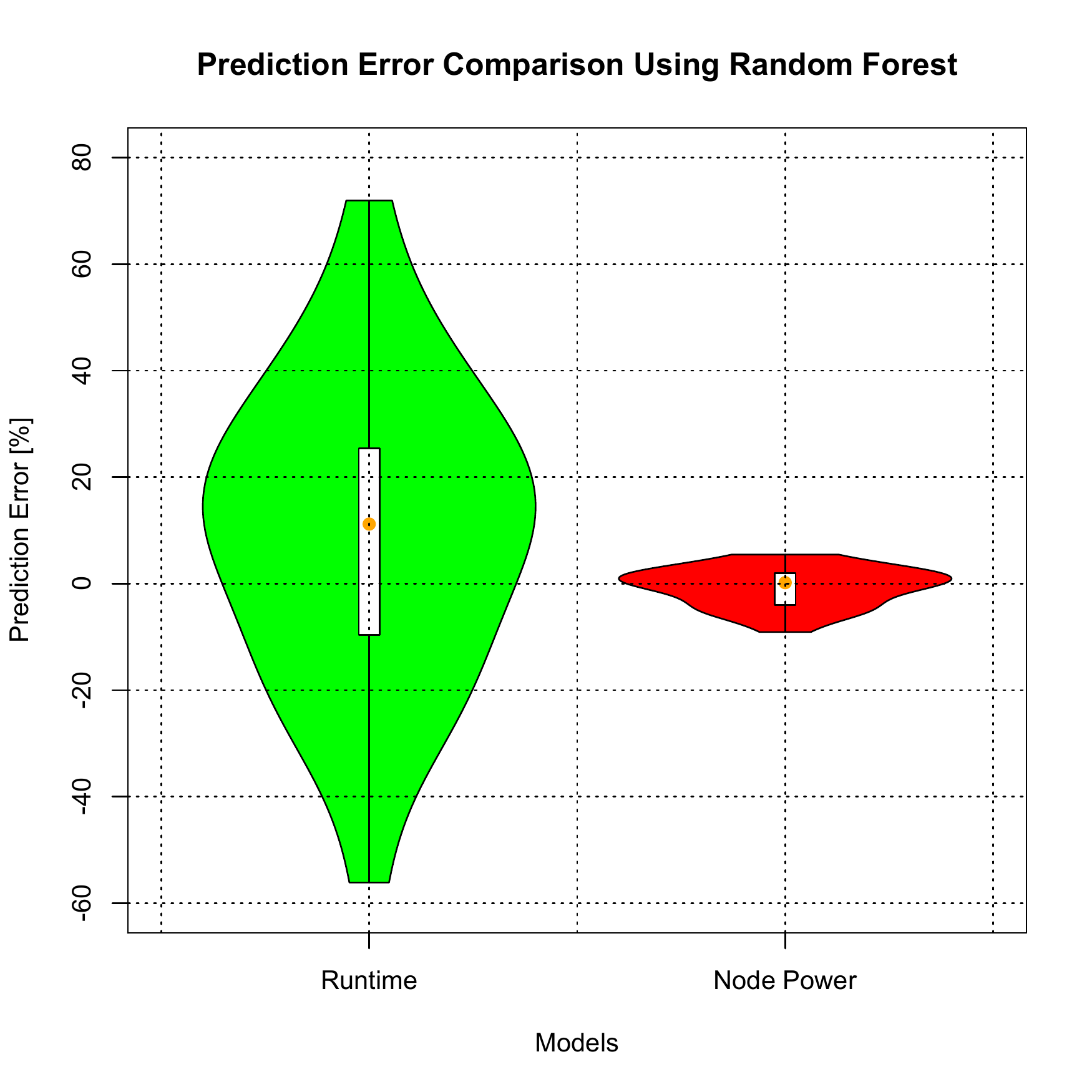}
 \caption{Prediction error rates using Random Forest}
\label{fig:44}       
\end{figure}

\begin{figure}
\center
 \includegraphics[height=2in, width=3in]{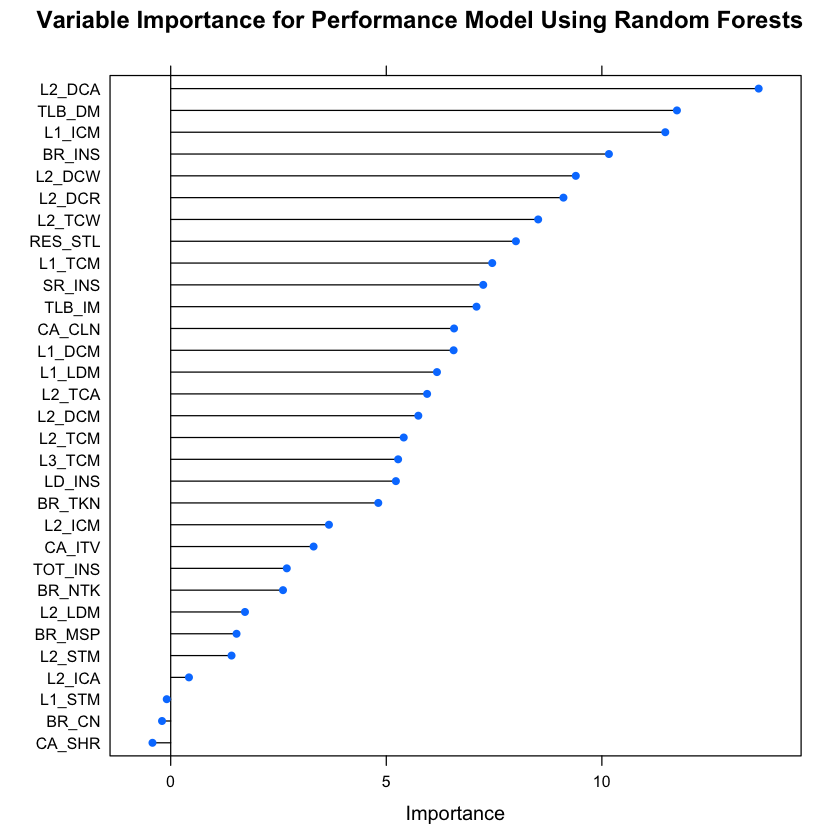}
 \caption{Variable importance for performance model of FTLA}
\label{fig:64}       
\end{figure}

\begin{figure}
\center
 \includegraphics[height=2in, width=3in]{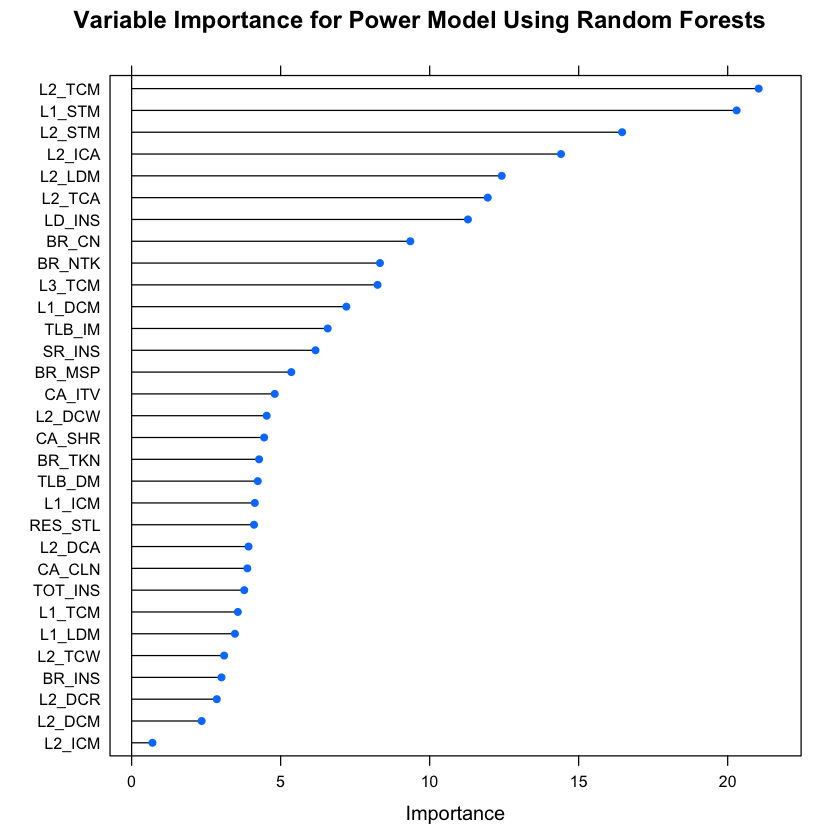}
 \caption{Variable importance for node power model of FTLA}
\label{fig:65}       
\end{figure}

Figure ~\ref{fig:44} shows the prediction error rates for runtime and node power models using Random Forests. The prediction error rates are between -56.15\% and 71.97\% in runtime, and between -9.09\% and 5.46\% in node power. The mean error rate is 9.48\% in runtime and -0.68\% in node power.  Figure ~\ref{fig:64} shows the variable importance for performance model of FTLA. Figure ~\ref{fig:65} shows the variable importance for node power model of FTLA. We observe that the top 3 counters in performance model are L2\_DCA, TLB\_DM, and L1\_ICM; the top 3 counters in power model are L2\_TCM, L1\_STM, and L2\_STM. Overall, L2 cache, TLB, and L1 cache mainly impact the performance and power by using Random Forests, however, it is interesting to observe that the top 3 counters in performance model are in the bottom of the counter list in power model, and the top 3 counters in power model are in the bottom of the counter list in performance model. The variable importance in random forest models is similar to that in cubist models, albeit in different order, because both are tree/rule-based models. 

\begin{figure}
\center
 \includegraphics[height=2in, width=3in]{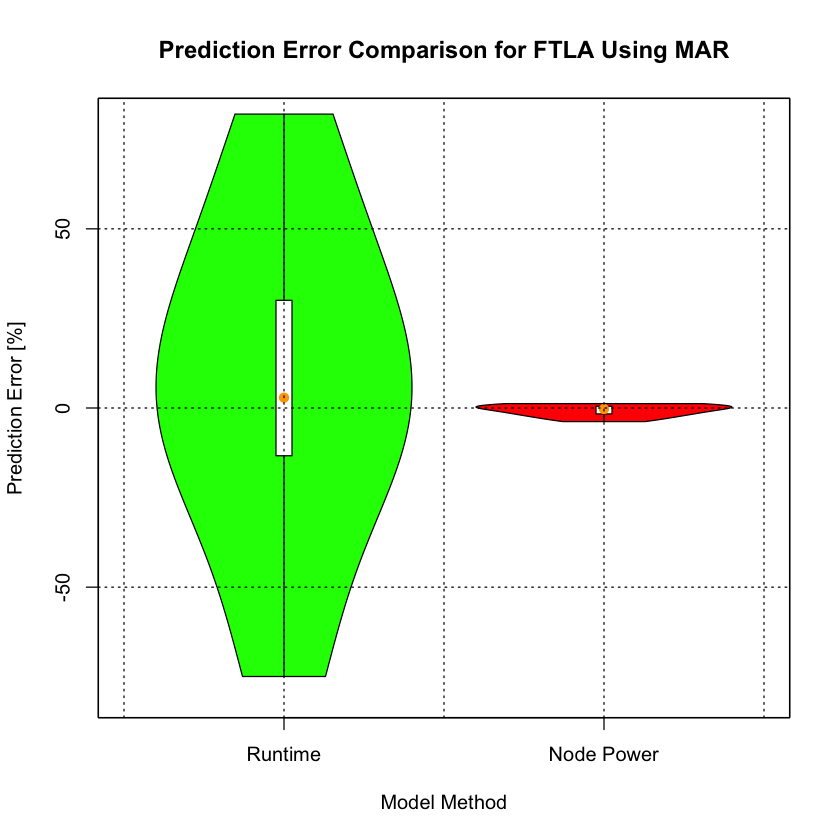}
 \caption{Prediction error rates using MAR}
\label{fig:82}       
\end{figure}

\begin{figure}
\center
 \includegraphics[height=2in, width=3in]{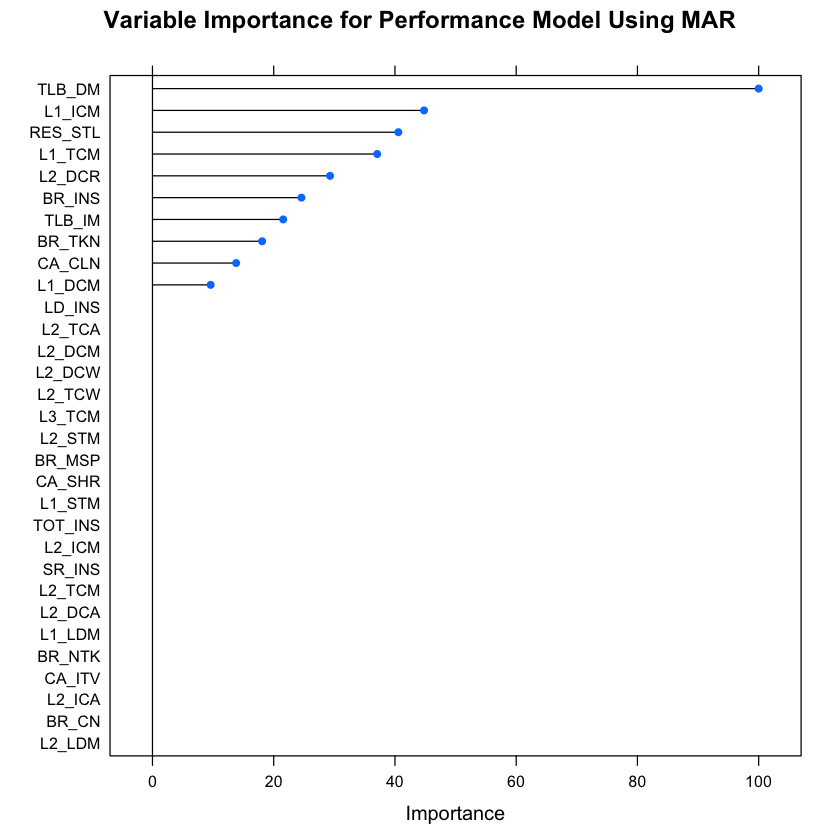}
 \caption{Variable importance for performance model of FTLA}
\label{fig:66}       
\end{figure}

\begin{figure}
\center
 \includegraphics[height=2in, width=3in]{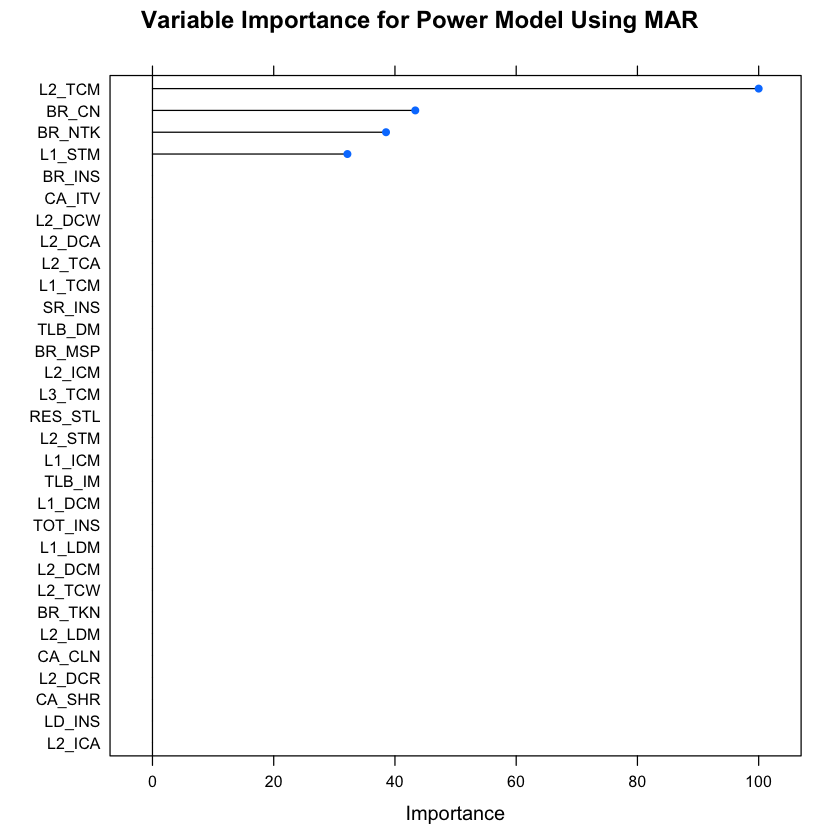}
 \caption{Variable importance for node power model of FTLA}
\label{fig:67}       
\end{figure}

Figure ~\ref{fig:82} shows the prediction error rates using Multivariate Adaptive Regression Spline are between -56.15\% and 71.97\% in runtime, and between -9.09\% and 5.46\% in node power. The average error rate is 9.48\% in runtime and -0.68\% in node power.  Figure ~\ref{fig:66} shows the variable importance with only 10 counters used in performance model of FTLA. Figure ~\ref{fig:67} shows the variable importance with only 4 counters used in node power model of FTLA. We observe that the top 3 counters in performance model are TLB\_DM, L1\_ICM, and RES\_STL; the top 3 counters in power model are L2\_TCM, BR\_CN, and BR\_NTK. Overall, TLB and L2 cache mainly impact the performance and power by using Multivariate Adaptive Regression Spline.  

In summary, for the four ML methods, we find that TLB\_DM is one of the dominant factors in performance models, and L2\_TCM is the dominant factor in power models.  This validates that L2\_TCM is the dominant factor in power models using MuMMI in Figure \ref{fig:23}. 
Given the datasets, 10 ML models have been fit to the datasets. Since the ML methods have their own way of learning the relationship between the predictors and the target object and provide different variable importance, it is hard to identify which ML provides the robust variable importance.

\subsection{HDC}
For HDC with the checkpointing file size of 2MB per MPI process (weak scaling), we ran the HDC with ten different four-levels checkpointing configurations on eight distinct numbers of cores (32, 64, 128,  256, 512, 640, 960, and 1024) with the CPU frequency of 2.3 GHz to collect the total 80 data samples. Each data sample includes 54 variables such as application name, system name, number of cores, number of iterations, checkpointing file size, Level 1 checkpoint, Level 2 checkpoint, Level 3 checkpoint, Level 4 checkpoint, CPU frequency, 32 available performance counters, runtime, system power, CPU power, memory power, and so on. We split the data as training and test datasets with the 80/20\% rule so that the training dataset consists of 64 samples, and the test dataset consists of 16 samples. For the fair comparison, we apply the same training and test datasets to all modeling methods.

\begin{figure}
\center
 \includegraphics[height=2in, width=3in]{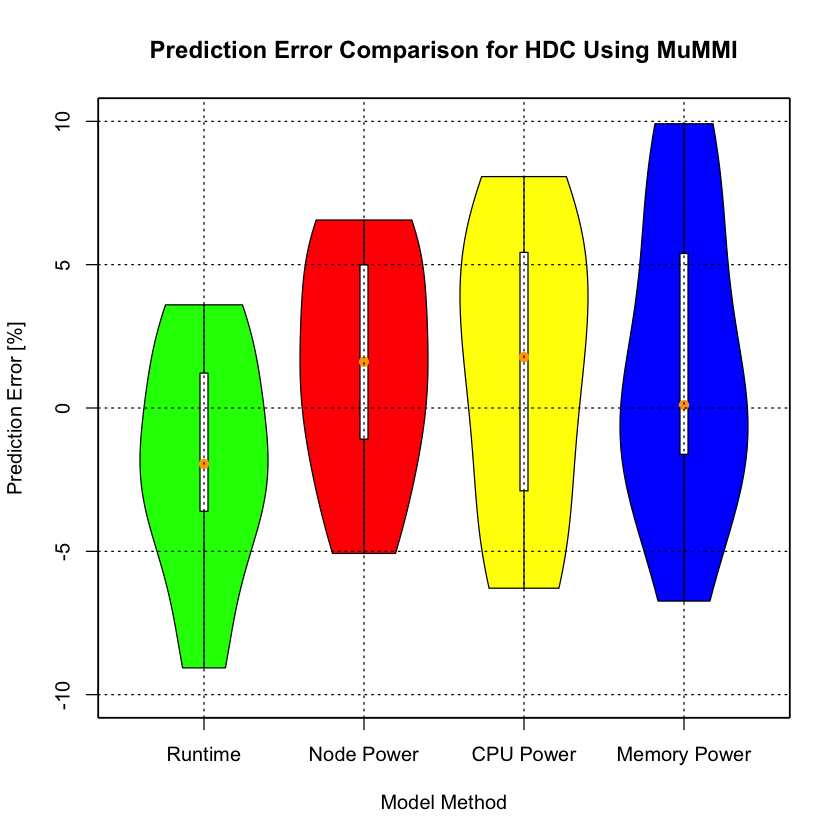}
 \caption{Prediction error rates for HDC using MuMMI}
\label{fig:50}       
\end{figure}

Figure ~\ref{fig:50} shows the prediction error rates for the models in runtime, node power, CPU power, and memory power using MuMMI. The prediction error rates are between -9.07\% and 3.60\% in runtime and between -5.07\% and 6.56\% in node power. The mean error rate is -1.88\% in runtime, 1.40\% in node power, 1.28\% in CPU power, and 1.55\% in memory power. Overall, this prediction is very accurate in runtime and power using MuMMI.

\begin{figure}
\center
 \includegraphics[height=2.4in, width=3.3in]{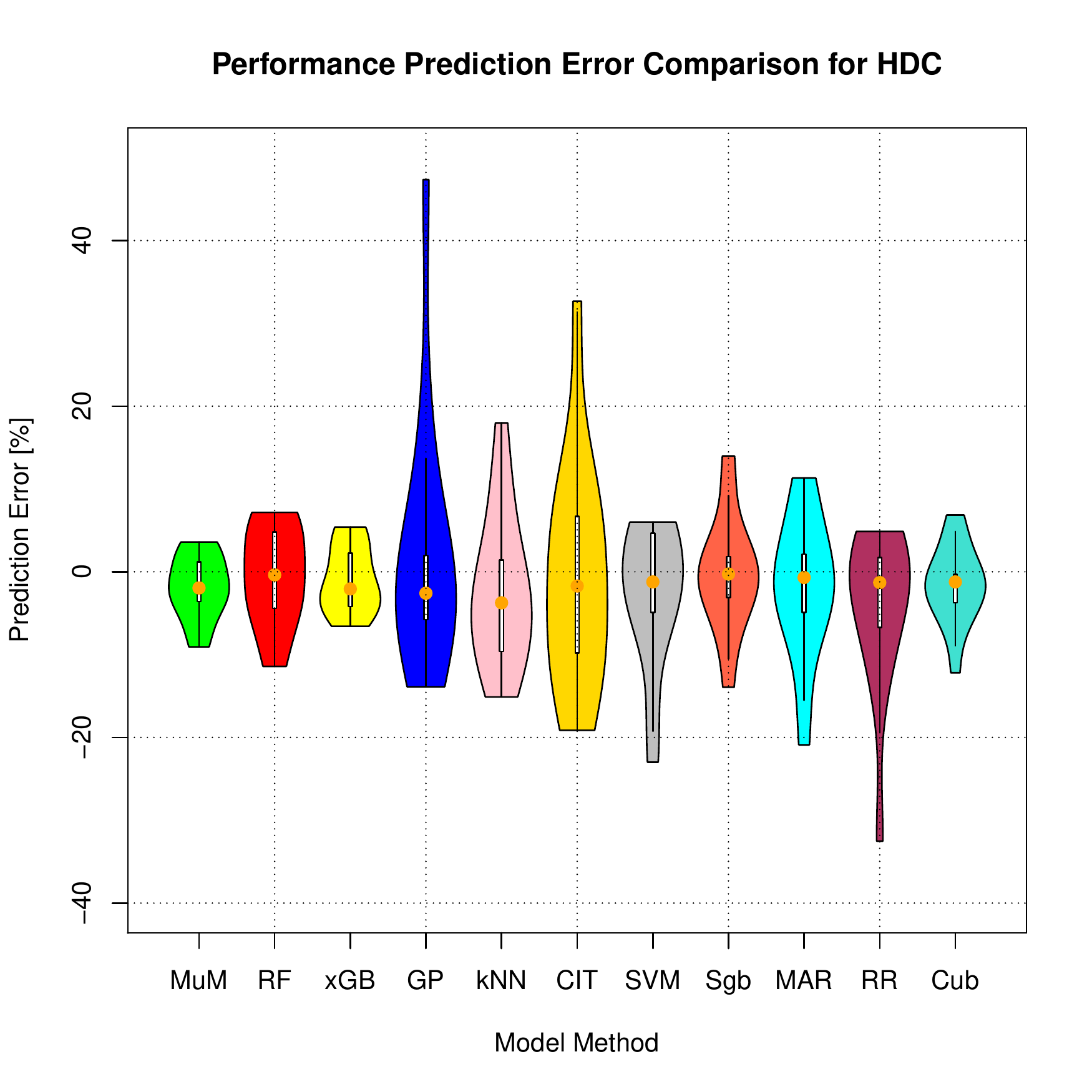}
 \caption{Prediction error rates (Runtime) for HDC }
\label{fig:51}       
\end{figure}

Figure \ref{fig:51} shows the performance prediction error rates using 10 ML methods and MuMMI (MuM). Based on these error rates, we observe that eXtreme Gradient Boosting (xGB) resulted in the lowest error rates in performance models among 10 ML methods, and for other ML methods, the maximum error rates are more than 11\%. Therefore, MuMMI outperformed all of them in performance prediction.    

\begin{figure}
\center
 \includegraphics[height=2.4in, width=3.3in]{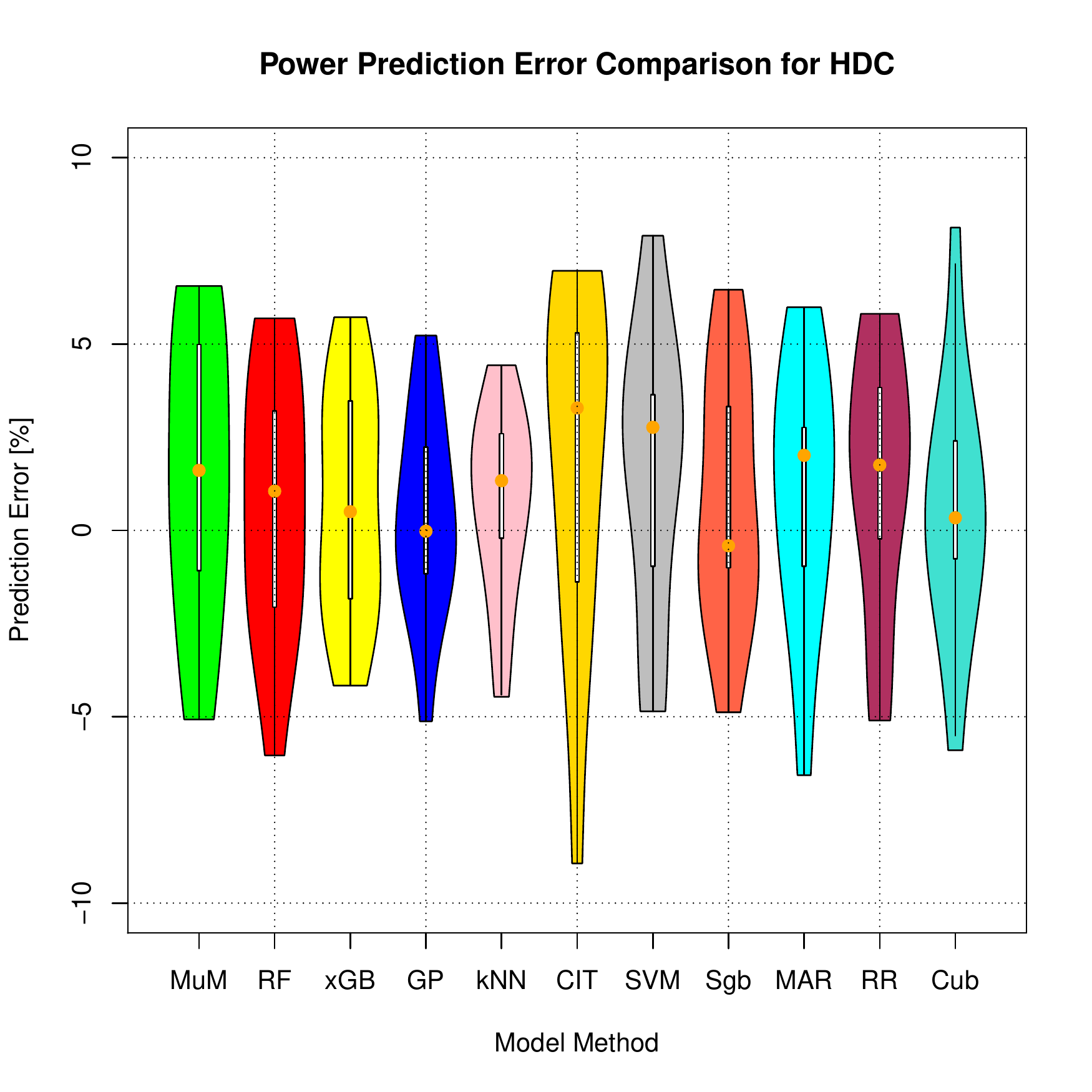}
 \caption{Prediction error rates (Node Power) for HDC }
\label{fig:52}       
\end{figure}

Figure \ref{fig:52} shows the node power prediction error rates using MuMMI and 10 ML methods. Based on these error rates, we observe that they are between -9\% and 9\%. k-Nearest Neighbors (kNN) and eXtreme Gradient Boosting (xGB) resulted in the lowest error rates in node power among 10 ML methods and outperformed MuMMI although the node power prediction error rates using MuMMI are between -5.07\% and 6.56\%.

\begin{figure}
\center
 \includegraphics[height=2in, width=3in]{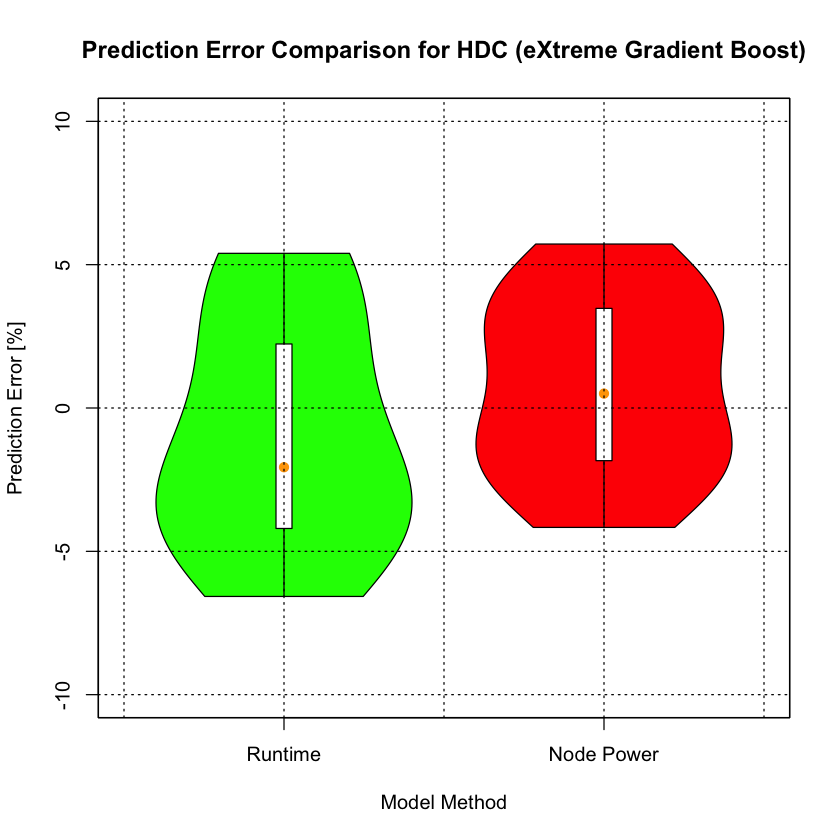}
 \caption{Prediction error rates using eXtreme Gradient Boosting}
\label{fig:78}       
\end{figure}

\begin{figure}
\center
 \includegraphics[height=2in, width=3in]{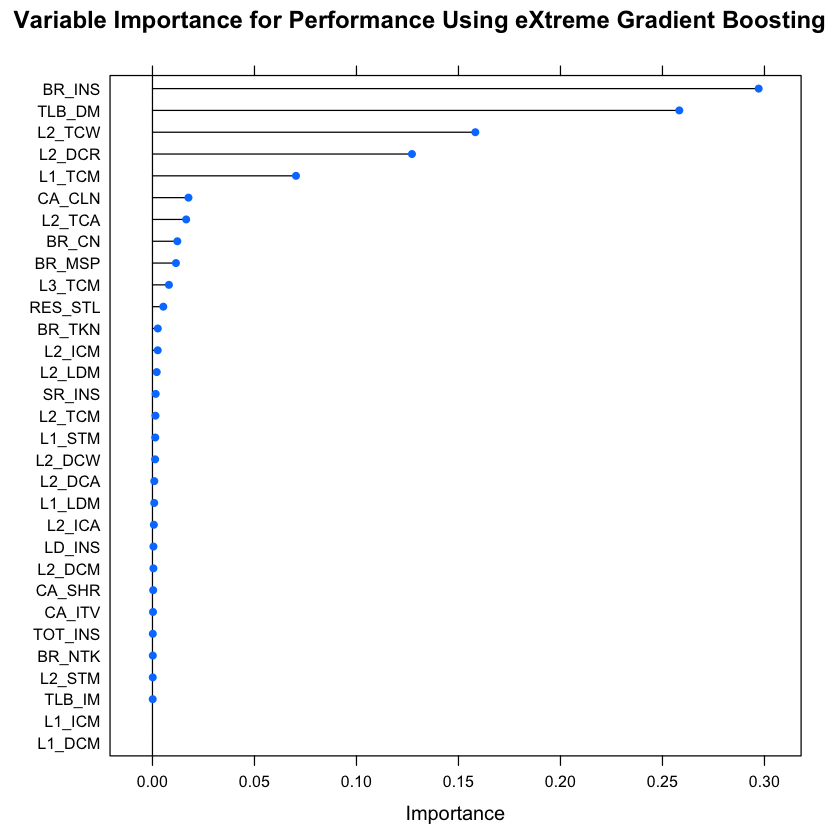}
 \caption{Variable importance for performance model of HDC}
\label{fig:70}       
\end{figure}

\begin{figure}
\center
 \includegraphics[height=2in, width=3in]{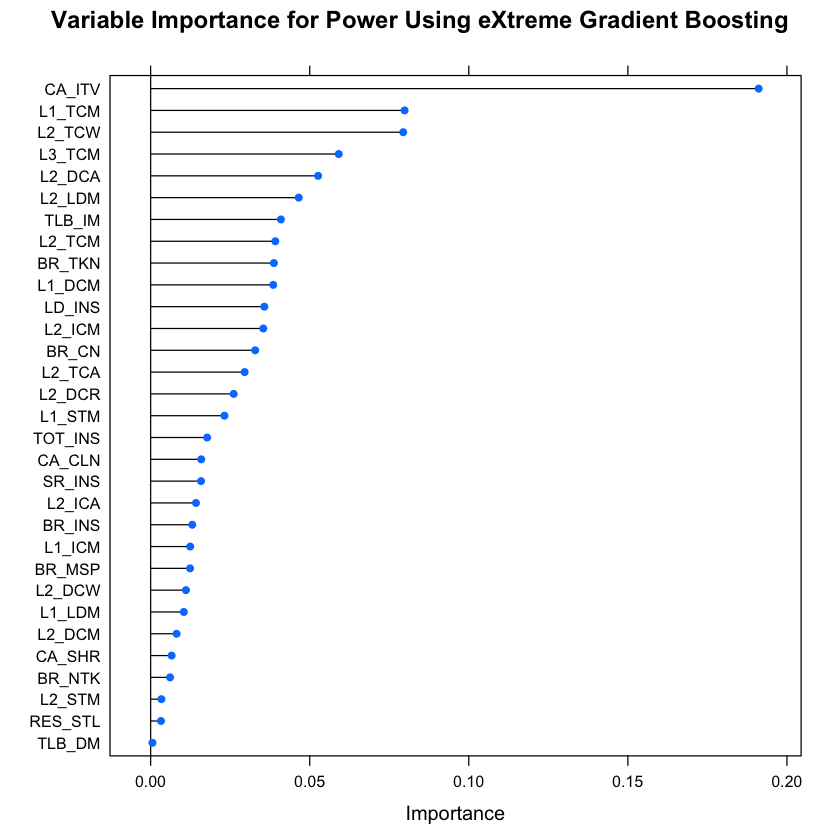}
 \caption{Variable importance for node power model of HDC}
\label{fig:71}       
\end{figure}

Figure ~\ref{fig:78} shows that the prediction error rates using eXtreme Gradient Boosting are between -6.57\% and 5.40\% in runtime, and between -4.17\% and 5.72\% in node power. The mean error rate is -0.92\% in runtime and 0.71\% in node power.  Figure ~\ref{fig:70} shows the variable importance for performance model of HDC. Figure ~\ref{fig:71} shows the variable importance for node power model of HDC. We observe that the top 3 counters in performance model are BR\_INS, TLB\_DM, and L2\_TCW; the top 3 counters in power model are CA\_ITV, L1\_TCM, and L2\_TCW. It is interesting to observe that the top counter  BR\_INS in performance model is in the bottom of the counter list in power model, and the top counter CA\_ITV in power model are in the bottom of the counter list in performance model.

\begin{figure}
\center
 \includegraphics[height=2in, width=3in]{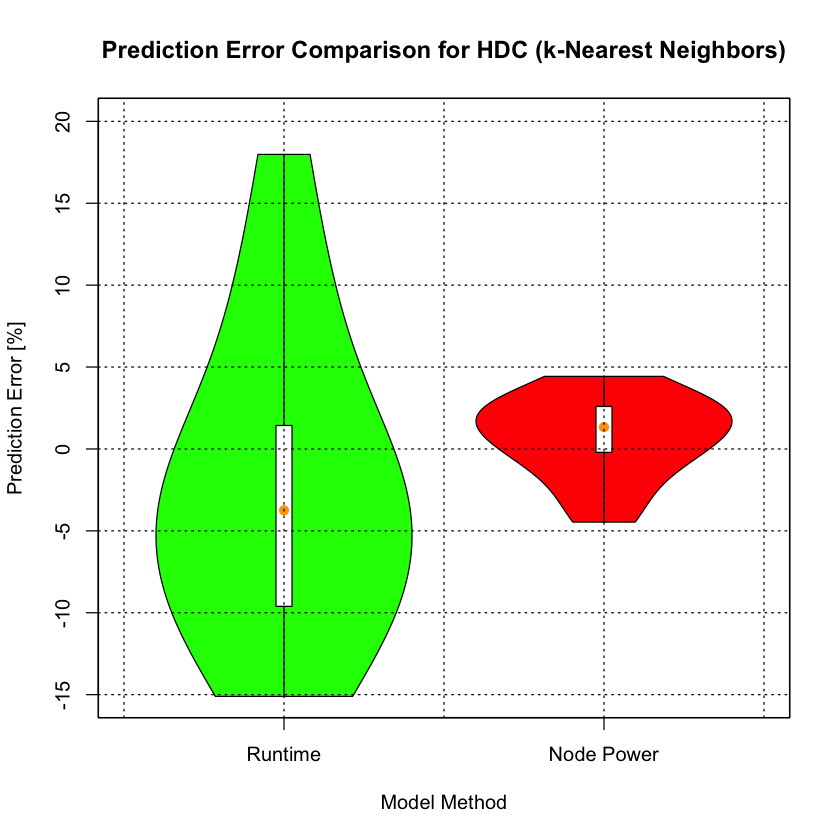}
 \caption{Prediction error rates using k-Nearest Neighbors}
\label{fig:81}       
\end{figure}

\begin{figure}
\center
 \includegraphics[height=2in, width=3in]{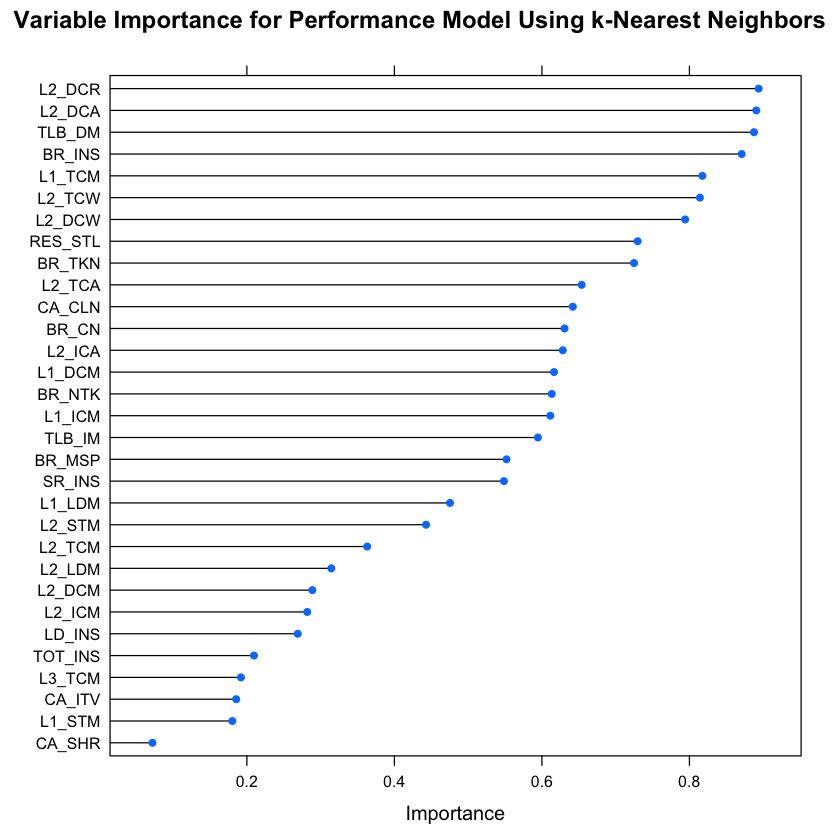}
 \caption{Variable importance for performance model of HDC}
\label{fig:76}       
\end{figure}

\begin{figure}
\center
 \includegraphics[height=2in, width=3in]{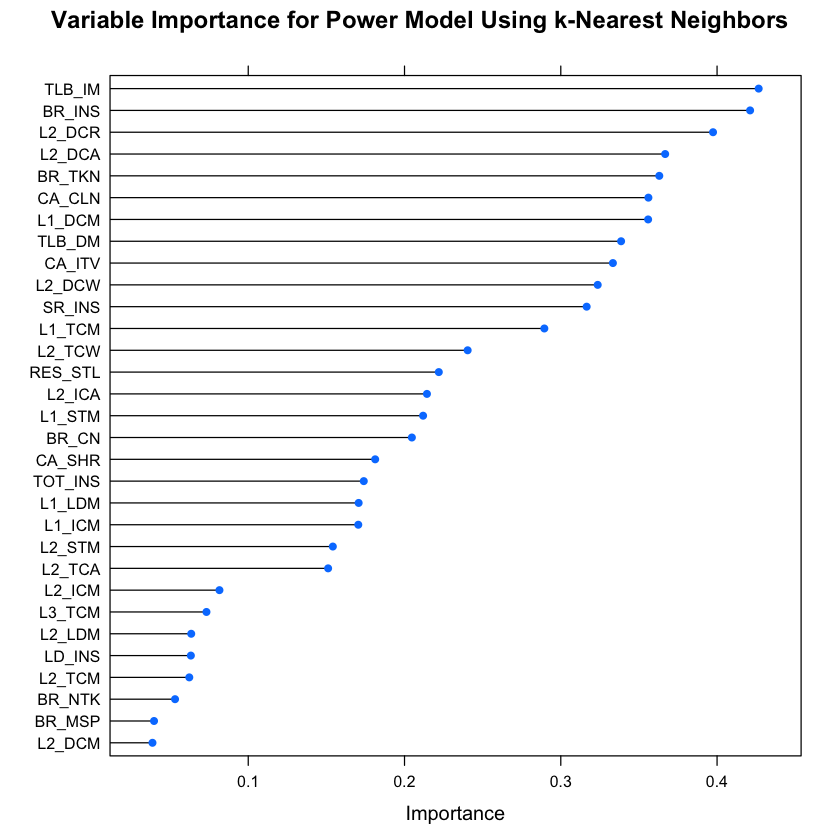}
 \caption{Variable importance for node power model of HDC}
\label{fig:77}       
\end{figure}

Figure ~\ref{fig:81} shows the prediction error rates using k-Nearest Neighbors are between -15.11\% and 17.98\% in runtime, and between -4.47\% and 4.43\% in node power. The mean error rate is -2.21\% in runtime and 0.71\% in node power.  Figure ~\ref{fig:76} shows the variable importance for performance model of HDC. Figure ~\ref{fig:77} shows the variable importance for node power model of HDC. We observe that the top 3 counters in performance model are L2\_DCR, L2\_DCA, and TLB\_DM; the top 3 counters in power model are TLB\_IM, BR\_CN, and L2\_DCR. Contrasting the importance results to eXtreme Gradient Boosting in Figure ~\ref{fig:70} and \ref{fig:71}, we see that 4 of the top 5 counters are the same in performance model and only 1 of the top 5 counters is the same in power model, and the importance orderings are much different.

\section{Conclusions}

 In this paper, we used MuMMI and 10 ML methods to model, predict and compare the performance and power of the FTLA and HDC. Our experiment results show that the prediction error rates in performance and power using MuMMI are less than 10\% for most cases. Based on the performance counters of these models, we identified the most significant performance counters for potential optimization efforts associated with the application characteristics on these systems, and we used our what-if prediction system to predict the theoretical performance and power of a possible application optimization. These performance and power models were generated from different system configurations and problem sizes, thus providing a broader understanding of the application's usage of the underlying architectures. This in turn resulted in more knowledge about the application's energy consumption on a given architecture. 
 
When we compare the prediction accuracy using MuMMI with that using 10 ML methods, we observe that MuMMI resulted in more accurate prediction in both performance and power. Since the 10 ML methods have their own way of learning the relationship between the predictors and the target object and provide different variable importance, it is hard to identify which ML provides the robust variable importance for potential improvements. To address the issue in our future work, we plan to utilize ensemble learning to combine several ML methods to result in more accurate models and provide the robust variable importance for the latent improvements. Performance and power modeling tools like MuMMI is able to aid in application optimizations for energy efficiency, power or energy-aware job schedulers, and system performance and power tuning. The general methodology presented in this paper can be applied to large scale scientific applications \cite{WT16} and deep learning applications \cite{WT19} on other parallel systems.


\section*{Acknowledgments}
This work was supported in part by Laboratory Directed Research and Development (LDRD) funding from Argonne National Laboratory, provided by the Director, Office of Science, of the U.S. Department of Energy under contract DE-AC02-06CH11357, and in part by NSF grants CCF-1801856. We acknowledge Argonne Leadership Computing Facility for use of Cray XC40 Theta and BlueGene/Q Mira under the DOE INCITE project PEACES and ALCF project EE-ECP, and Sandia National Laboratories for use of Intel Haswell Shepard testbed.

\end{document}